\documentclass[10pt,twocolumn,letterpaper]{article}

\usepackage{cvpr}              
\definecolor{cvprblue}{rgb}{0.21,0.49,0.74}
\usepackage[pagebackref,breaklinks,colorlinks,allcolors=cvprblue]{hyperref}
\usepackage{multicol}
\usepackage{multirow}
\usepackage{amsmath}
\usepackage{amssymb}
\usepackage{algorithm}
\usepackage{algorithmic}
\usepackage{graphicx}
\newtheorem{definition}{Definition}
\newtheorem{proposition}{Proposition}

\newtheorem{assumption}{Assumption}
\newtheorem{corollary}{Corollary}
\usepackage{subcaption}
\usepackage[accsupp]{axessibility}

\def\paperID{} 
\def\confName{CVPR}
\def\confYear{2026}

\title{When Robots Obey the Patch: Universal Transferable Patch Attacks on Vision-Language-Action Models}

\author{Hui Lu$^{1,2}$ \quad
Yi Yu$^{2}$\thanks{Corresponding author.} \quad
Yiming Yang$^{3}$ \quad
Chenyu Yi$^{2}$ \quad
Qixin Zhang$^{3}$ \quad
Bingquan Shen$^{4}$ \quad \\
Alex C.~Kot$^{2}$\quad 
Xudong Jiang$^{2}$ \\
$^{1}$ROSE Lab, Interdisciplinary Graduate Programme, Nanyang Technological University\\
$^{2}$ROSE Lab, School of Electrical and Electronic Engineering, Nanyang Technological University\\
$^{3}$CCDS, Nanyang Technological University \quad
$^{4}$DSO National Laboratories\\
{\tt\small \{hui007,\!\! yu\!.\!yi,\!\! yiming014,\!\! cyyi,\!\! qixin\!.\!zhang,\!\! eackot,\!\! exdjiang\}@ntu.edu.sg, sbingqua@dso.org.sg}
}

\begin{document}
\maketitle
\begin{abstract}
Vision-Language-Action (VLA) models are vulnerable to adversarial attacks, yet universal and transferable attacks remain underexplored, as most existing patches overfit to a single model and fail in black-box settings. To address this gap, we present a systematic study of \textbf{universal, transferable adversarial patches} against VLA-driven robots under unknown architectures, finetuned variants, and sim-to-real shifts. We introduce \textbf{UPA-RFAS} (Universal Patch Attack via Robust Feature, Attention, and Semantics), a unified framework that learns a single physical patch in a shared feature space while promoting cross-model transfer. UPA-RFAS combines (i) a feature-space objective with an $\ell_1$ deviation prior and repulsive InfoNCE loss to induce transferable representation shifts, (ii) a robustness-augmented two-phase min-max procedure where an inner loop learns invisible sample-wise perturbations and an outer loop optimizes the universal patch against this hardened neighborhood, and (iii) two VLA-specific losses: Patch Attention Dominance to hijack text$\to$vision attention and Patch Semantic Misalignment to induce image-text mismatch without labels. Experiments across diverse VLA models, manipulation suites, and physical executions show that UPA-RFAS consistently transfers across models, tasks, and viewpoints, exposing a practical patch-based attack surface and establishing a strong baseline for future defenses.
Code is at \url{https://github.com/yuyi-sd/UPA-RFAS}.
\end{abstract}    
\section{Introduction}
\label{sec:intro}

Vision-Language-Action (VLA) models have made significant strides, facilitating open-world manipulation \citep{black2024pi_0, black2024pi05}, language-conditioned planning \citep{kim2024openvla}, and cross-embodiment transfer \citep{brohan2022rt, zitkovich2023rt}. By coupling a visual encoder with language grounding and an action head, modern VLA models are capable of parsing free-form instructions and executing multi-step skills in both simulation and the physical world \citep{liu2023libero, walke2023bridgedata}. Despite their potential, such multi-modal pipelines are vulnerable to structured visual perturbations, \textit{aka} adversarial attacks \citep{brown2017adversarial,li2025pbcat, eykholt2018robust,zhao2023fast,zhao2024catastrophic,zhao2024adversarial,zhao2024survey,xia2024mitigating,yu2025towards,zhou2023advclip,zhou2024darksam,zhou2024securely,zhou2025numbod,zhou2025sam2,zhou2025badvla,DBLP:journals/tifs/LiLWTZ25,li2024dat,li2026aegis}, which can mislead perception, disrupt cross-modal alignment, and cascade into unsafe actions. This issue is particularly severe in robotics, as attacks that merely flip a class in perception can translate into performance drops, collisions, or violations of task constraints on real-world systems \citep{zhang2024badrobot,wang2025exploring, robey2025jailbreaking}. Motivated by that, we conduct a systematic study of universal and transferable adversarial patches for VLA-driven robots, where black-box conditions, varying camera poses, and domain shifts from simulation to the real world are the norm in practical robotic deployments. 

Though vulnerabilities in VLAs have received growing attention \citep{wang2025exploring,xu2025model,fei2025libero, zhou2025badvla, robey2025jailbreaking}, universal and transferable attacks remain largely under-explored. Reported patches often co-adapt to a specific model, datasets, or prompt template, and their success degrades sharply on unseen architectures or finetuned variants \citep{kim2025fine}, precisely the black-box regimes that matter for safety assessment. As a result, current evaluations can overestimate security when the attacker lacks white-box access, and underestimate the risks of patch-based threats that exploit cross-modal bottlenecks \cite{jia2025adversarial}. Bridging this gap requires attacks that generalize across families of VLAs (\textit{e.g.,} OpenVLA \citep{kim2024openvla}, lightweight OFT variants \citep{kim2025fine}, and flow-based policies such as $\pi_o$ \cite{black2024pi_0}).

\begin{figure*}
    \centering
    \includegraphics[width=1.0\linewidth]{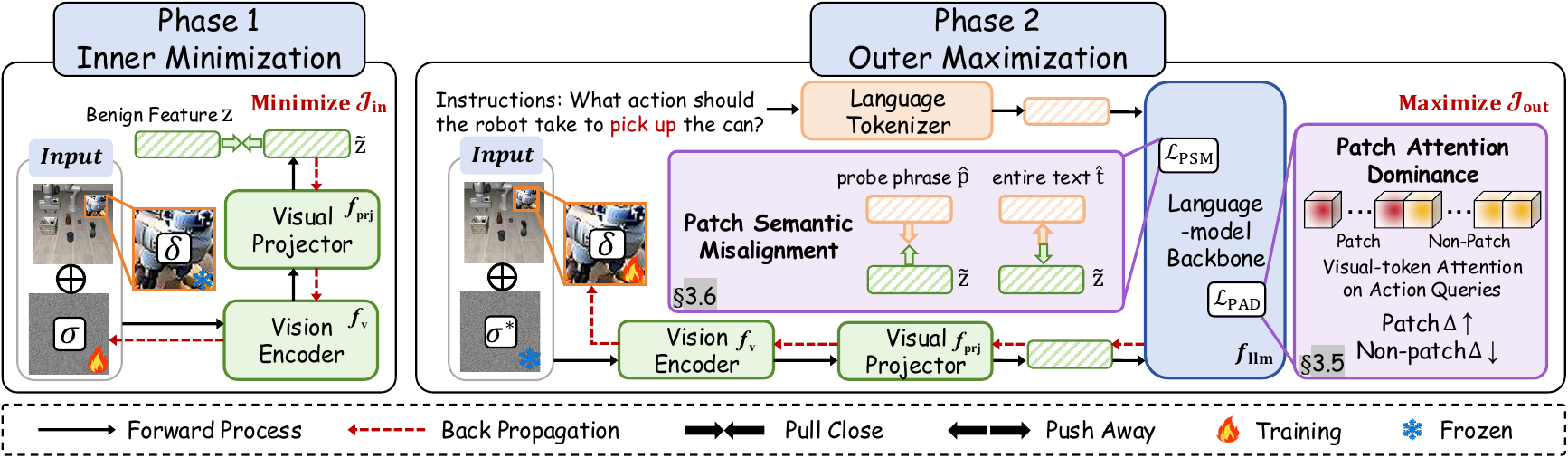}
    \vspace{-5mm}\caption{\textbf{Overall transferable patch attack (UPA-RFAS) for VLA robotics. }
The framework operates in two coordinated stages within a shared feature-space objective.
\emph{Phase 1 – Inner minimization} learns a small, invisible, sample-wise perturbation $\boldsymbol{\sigma}$ via PGD that \emph{minimizes} the feature objective $\mathcal{J}_{\mathrm{in}}$ (§~\ref{sec:base}) with the patch frozen (§~\ref{sec:dual}).
\emph{Phase 2 – Outer maximization} freezes $\boldsymbol{\sigma}$ and optimizes a \emph{single} physical patch $\boldsymbol{\delta}$ to \emph{maximize} $\mathcal{J}_{\mathrm{out}}$ (§~\ref{sec:alg}), which combines an $\ell_1$ deviation with a repulsive contrastive term and two VLA-specific objectives: \textbf{Patch Attention Dominance} (PAD) (§~\ref{sec:attn}) and \textbf{Patch Semantic Misalignment} (PSM) (§~\ref{sec:text}).
\textcolor{red}{Red} dashed arrows indicate back-propagation.
UPA-RFAS yields a universal physical patch that transfers across models, prompts, and viewpoints.}
    \label{fig:placeholder}
    \vspace{-2mm}
\end{figure*}

We bridge the surrogate and victim gap by learning a universal patch in a shared feature space, guided by two principles: enlarge surrogate-side deviations that provably persist on the target, and concentrate changes along stable directions. An $\ell_1$ deviation term drives sparse, high-salience shifts \citep{chen2018ead} that avoid surrogate-specific quirks, while a repulsive InfoNCE loss \citep{chen2020simple} pushes patched features away from their clean anchors along batch-consistent, high-CCA directions \citep{raghu2017svcca}, strengthening black-box transfer.
To further raise universality, we adopt a {Robustness-augmented Universal Patch Attack (RAUP)}. The inner minimization loop learns a small, sample-wise invisible perturbation that reduces the feature-space objective around each input, emulating local adversarial training and hardening the surrogate. The outer maximization loop then optimizes a single physical patch against this hardened neighborhood with randomized placements and transformations, distilling the stable, cross-input directions revealed by the inner loop.
For robotics, we further couple feature transfer with policy-relevant signals: \emph{(i)} the {Patch Attention Dominance (PAD)} loss increases patch-routed text$\to$vision attention and suppresses non-patch increments with a one-sided margin, yielding location-agnostic attention attraction; \emph{(ii)} the {Patch Semantic Misalignment (PSM)} loss pulls the pooled patch representation toward probe-phrase anchors while repelling it from the current instruction embedding, creating a persistent image–text mismatch that perturbs instruction-conditioned policies without labels. Together, these components form \textbf{Universal Patch Attack via Robust Feature, Attention, and Semantics (UPA-RFAS)}, a universal, transferable patch framework that aligns attack feature shifts, cross-modal attention, and semantic steering.



Our contributions are summarized as follows:
\begin{itemize}
    \item We present the first \emph{universal, transferable} patch attack framework for VLA robotics, using a feature-space objective that combines $\ell_1$ deviation with repulsive contrastive alignment for model-agnostic transfer.
    \item We propose a \emph{robustness-augmented} universal patch attack, with invisible sample-wise perturbations as hard augmenters and a universal patch trained under heavy geometric randomization.
    \item We design two VLA-specific losses: \emph{Patch Attention Dominance} and \emph{Patch Semantic Misalignment} to hijack text$\to$vision attention and misground instructions.
    \item Extensive experiments across VLA models, tasks, and sim-to-real settings show strong black-box transfer, revealing a practical patch-based threat and a transferable baseline for future defenses.
\end{itemize}

\vspace{-2mm}
\section{Related Work}
\vspace{-2mm}
\textbf{Vision-Language-Action (VLA) Models.}
Advances in large vision–language models (LVLMs) \citep{beyer2024paligemma, steiner2024paligemma, zhai2023sigmoid,zhu2023languagebind,oquab2023dinov2} have prompted robotic manipulation to leverage the powerful capabilities of vision–language modeling. VLA models extend LVLMs to robotic control by coupling perception, language grounding, and action generation. Autoregressive VLAs discretize actions into tokens and learn end-to-end policies from large demonstrations, yielding scalable instruction-conditioned manipulation \citep{zitkovich2023rt,kim2024openvla,li2024manipllm, wen2025tinyvla, brohan2022rt, pertsch2025fast}. Diffusion-based VLAs generate continuous trajectories with denoisers for smooth rollouts and flexible conditioning, at the cost of higher inference latency \citep{black2024pi_0,black2024pi05,bjorck2025gr00t,li2024cogact,wen2025diffusionvla}. RL-enhanced VLAs optimize robustness and adaptability beyond supervised imitation by introducing reinforcement objectives over VLA backbones \citep{tan2025interactive, lu2025vla, guo2025improving}. VLA models exemplify strong vision–language alignment for compositional task understanding and end-to-end action generation, while raising new questions about robustness under instruction-conditioned deployment.

\noindent\textbf{Adversarial Attacks in Robotics.}
Adversarial attacks are commonly grouped by access level: white-box methods assume full knowledge and directly use model gradients \cite{goodfellow2014explaining,yu2022towards,wang2024benchmarking,yu2026spikeretiming}, whereas black-box methods operate without internals-either by querying the model for feedback \cite{chen2017zoo,liu2024difattack,liu2024difattack++} or by exploiting cross-model transferability of crafted examples~\cite{lu2026pretrain,lu2026make,xia2024transferable}. To strengthen transfer, optimization-driven approaches refine or stabilize gradient signals to avoid local minima arising from mismatched decision boundaries across architectures \cite{liu2016delving,kurakin2018adversarial,dong2018boosting,lin2019nesterov,wang2021enhancing,zhu2023boosting}. Augmentation-based strategies diversify inputs to induce gradient variation and reduce overfitting to a single surrogate \cite{xie2019improving,lin2019nesterov,dong2019evading,wang2021admix,wang2024boosting,l2t}. Finally, feature-space attacks aim at intermediate representations to promote cross-model invariance and further improve transfer \cite{fda, fia, naa, rpa}.
Patch-based physical attacks \citep{xu2020adversarial, shekhar2025adversarial, xiao2021improving, huang2023t, brown2017adversarial} are practical for real-world deployment, remaining effective under changes in viewpoint and illumination, which makes them suitable for robotic systems. VLA models \citep{team2024octo, huang2023voxposer} couple visual and linguistic modalities to align perception with action, and visual streams are high-dimensional and can be subtly perturbed in ways that are difficult to detect \citep{athalye2018obfuscated, tramer2022detecting}. Accordingly, our work targets the visual modality with a universal, transferable patch attack. To our knowledge, it is the first to investigate black-box transfer vulnerabilities of VLAs in real-world settings.

\section{Methodology}
\subsection{Preliminary}
\noindent\textbf{Adversarial Patch Attack.}
We consider a robot whose decisions are based on RGB visual streams
$\mathbf{x}_t \in [0,1]^{H \times W \times 3}$ across time step $t$.
An adversary tampers with this stream using a
\emph{single universal} patch
$\boldsymbol{\delta} \in [0,1]^{h_p \times w_p \times 3}$.
At each time step $t$, an area-preserving geometric transformation
$T_t \sim \mathcal{T}$ (\textit{e.g.,} random position, skew, and rotation)
is sampled, and the transformed patch is rendered onto the frame.
Given $\mathbf{M}_{T_t} \!\in\! \{0,1\}^{H \times W}$ as the binary placement
mask induced by $T_t$, and
$\mathcal{R}(\boldsymbol{\delta}; T_t) \in [0,1]^{H \times W \times 3}$
as the rendered patch,
the pasting function $\mathcal{P}$ and resulting patched frame is
\begin{equation}
\begin{split}
\tilde{\mathbf{x}}_t
&= \mathcal{P}(\mathbf x_t,\boldsymbol{\delta},T_t) \\
&= \big( \mathbf{1} - \mathbf{M}_{T_t} \big) \odot \mathbf{x}_t
  + \mathbf{M}_{T_t} \odot \mathcal{R}(\boldsymbol{\delta}; T_t)~~~\text{s.t.}\;\mathcal{S}(\boldsymbol{\delta}) < \rho,
\end{split}
\label{eq:applypatch-compact}
\end{equation}
where $\odot$ is the Hadamard product, $\mathbf{1}$ is an all-ones matrix,
$\mathcal{S}(\cdot)$ returns the patch area (\textit{i.e.,} $h_p \times w_p$), and $\rho$ is an area budget
controlling the maximal visible size of the patch.

Let $\pi$ denote a \emph{victim} policy.
Given visual inputs $\mathbf{x}$ drawn from a task distribution $p(\mathbf{x})$ and random
patch placements $T_t \sim \mathcal{T}$, an adversarial patch attack aims to
\emph{learn} a single universal patch $\boldsymbol{\delta}$ that maximizes
an evaluation objective $\mathcal{J}_{\mathrm{eval}}$ (\textit{e.g.,} task loss
increase or action-space deviation~\cite{wang2025exploring}) under pasting $\mathcal{P}$ in Eq.~\ref{eq:applypatch-compact} and these
randomized conditions:
\begin{equation}
\boldsymbol{\delta}^{\star}
\;\in\;
\arg\max_{\mathcal{S}(\boldsymbol{\delta}) < \rho}
\;\mathbb{E}_{\substack{\mathbf{x}\sim p(\mathbf{x})\\T_t\sim\mathcal{T}}}\!
\left[
  \mathcal{J}_{\mathrm{eval}}\Big(\mathcal{P}(\mathbf{x},\boldsymbol{\delta},T_t);\pi\Big)
\right].
\label{eq:universal-transfer}
\end{equation}
This objective captures a \emph{single} patch that is robustly effective
across time, viewpoints, and scene configurations.

\vspace{1mm}
\noindent\textbf{VLAs.}
We follow the OpenVLA formulation~\cite{kim2024openvla}, where a policy is decomposed into a
\emph{vision encoder} $f_{\mathrm{v}}$, a \emph{visual projector} $f_{\mathrm{prj}}$, and a
\emph{language-model backbone} $f_{\mathrm{llm}}$ equipped with an \emph{action head}
$f_{\mathrm{act}}$.
Given an RGB observation $\mathbf{x}$ and an instruction $c$, the model predicts
an action vector $\mathbf{y}$ as
\begin{equation}
\label{eq:openvla-forward}
\begin{aligned}
\mathbf{y}
\!=\!
\mathrm{OpenVLA}(\mathbf{x}, c) \!=\!
f_{\mathrm{act}}\!\left(
    f_{\mathrm{llm}}\!\big(
        [\, f_{\mathrm{prj}}(f_{\mathrm{v}}(\mathbf{x})),\mathrm{tok}(c)]
    \big)
\right).    
\end{aligned}
\end{equation}
The computation can be unpacked as:
\textbf{(i)} the vision encoder $f_{\mathrm{v}}$ maps the image into a set of multi-granularity
visual embeddings, for example by concatenating DINOv2 \citep{oquab2023dinov2} and SigLIP \citep{zhai2023sigmoid} features, yielding
$\mathbf{E}_v \in \mathbb{R}^{N_v \times D_v}$ from $\mathbf{x}$;
\textbf{(ii)} the projector $f_{\mathrm{prj}}$ aligns these embeddings to the LLM token space,
producing visual tokens $\mathbf{Z}_v \in \mathbb{R}^{N_v' \times D_t}$;
\textbf{(iii)} the backbone $f_{\mathrm{llm}}$ takes the concatenation of $\mathbf{Z}_v$ and the
tokenized command $\mathrm{tok}(c)$, and fuses them into hidden states
$\mathbf{H}_\ell$;
\textbf{(iv)} the action head $f_{\mathrm{act}}$ decodes $\mathbf{H}_\ell$ into the continuous
control output $\mathbf{y} \in \mathbb{R}^{D_a}$ (\textit{e.g.,} a 7-DoF command).



\subsection{Problem Formulation}
Existing VLA patch attacks \cite{wang2025exploring} assume white-box access to the victim model, which limits their
practicality and says little about cross-policy transfer.
In our setting, the attacker instead only has gradient access to a
\emph{single} surrogate model $\hat{\pi}$ and aims to learn one
universal patch that transfers to a family of unseen target policies $\Pi_{\mathrm{tgt}}$.
To formalize this threat model, we separate
\emph{optimization} and \emph{evaluation}: the patch is optimized in the
surrogate feature space via a differentiable objective
$\mathcal{J}_{\mathrm{tr}}$, and its success is assessed by an
evaluation objective $\mathcal{J}_{\mathrm{eval}}$ on target policies
drawn from $\Pi_{\mathrm{tgt}}$.
Following \citep{wang2025exploring}, we adopt the untargeted attack setting, and summarize this transferable patch attack as follows.

\begin{definition}
\textup{\textbf{(Transferable adversarial patch attack via VLA feature space)}}\label{def1}
Let $\hat{\pi}$ be a surrogate model and $\Pi_{\mathrm{tgt}}$ a family of target policies.
Let $f_{\hat{\pi}}(\cdot)$ extract
features from $\hat{\pi}$. 
A patch $\boldsymbol{\delta}$ is a \emph{universal transferable adversarial patch in the VLA feature space}, it satisfies
\begin{equation}
\label{eq:feature-def}
\begin{aligned}
&\max_{\ \boldsymbol{\delta}_s}\quad 
\mathbb{E}_{\pi\sim\Pi_{\mathrm{tgt}}}\mathbb{E}_{\substack{\mathbf{x}\sim p(\mathbf{x})\\T_t\sim\mathcal{T}}}\!
\left[
  \mathcal{J}_{\mathrm{eval}}\Big( \mathcal{P}(\mathbf{x},\boldsymbol{\delta}_s,T_t);\pi\Big)
\right]
 \\
 &\text{s.t.}\;\boldsymbol{\delta}_s\in 
\arg\max_{\boldsymbol{\delta}}
\;\mathbb{E}_{\substack{\mathbf{x}\sim p(\mathbf{x})\\T_t\sim\mathcal{T}}}\left[
  \mathcal{J}_{\mathrm{tr}}\!\Big(\mathcal{P}(\mathbf{x},\boldsymbol{\delta},T_t);\hat{\pi}\!\Big)
\right],
\end{aligned}
\end{equation}
where $\mathcal{J}_{\mathrm{tr}}$ measures feature discrepancy using $\Delta$:
\begin{equation}
\mathcal{J}_{\mathrm{tr}}\!\Big(\mathcal{P}(\mathbf{x},\boldsymbol{\delta},T_t);\hat{\pi}\Big)=\Delta\Big(f_{\hat{\pi}}\big(\mathcal{P}(\mathbf{x},\boldsymbol{\delta},T_t)\big), f_{\hat{\pi}}\big(\mathbf{x}\big)\Big).\label{eq:feature_measurement}
\end{equation}
\end{definition} 
Here, $\mathcal{P}$ is the pasting function defined in Eq.~\ref{eq:applypatch-compact}, and $\mathcal{J}_\mathrm{tr}$ is the transferable attack strategy. Although $\hat{\pi}$ and $\pi$ differ in training recipe and data, we probe whether their feature spaces admit a stable \emph{cross-model relation} as follows:


\vspace{1mm}
\noindent\textbf{Shared Representational Structure across VLA Policies.}
Empirically, we observe a strong linear relationship between the surrogate and
target feature spaces.
Let $\mathbf{z}_s$ and $\mathbf{z}_t$ denote visual features from $\hat{\pi}$
and $\pi$ on the same inputs.
We first apply Canonical Correlation Analysis (CCA) to test whether these
representations lie in a \emph{shared linear subspace}: large top Canonical
Correlation indicates a near-invertible linear map aligning the two
subspaces~\citep{raghu2017svcca,morcos2018insights}.
In parallel, we fit a \emph{linear regression probe} from $\mathbf{z}_s$ to
$\mathbf{z}_t$ and use the explained variance ($R^2$) to quantify how well a
\emph{single} linear map accounts for the target features, complementing CCA’s subspace view~\citep{alain2016understanding, kornblith2019similarity}.
In our case, $R^2 \!\approx\! 0.654$ together with near-unity top-$k$
Canonical Correlations indicates a shared low-dimensional subspace, with some
residual components not captured by one linear map.
Consequently, patch updates that steer $\hat{\pi}$’s features within this
shared subspace tend to induce homologous displacements in $\pi$, supporting the transferability of patches.
Motivated by these observations, we make the following Assumption~\ref{asm:lin}.

\subsection{Learning Transferable Patches with Feature-space $\ell_1$ and Contrastive Misalignment}
\label{sec:base}

Let $f_{\hat{\pi}}, f_{\pi}\!:\!\mathcal{X}\!\!\to\!\!\mathbb{R}^d$ be the surrogate and target encoders with dimension $d$, where $f_\cdot$ consists of vision encoder $f_\mathrm{v}$ and visual projector $f_\mathrm{prj}$.
For any pair $(\mathbf{x}_i,\tilde{\mathbf{x}}_i)$, define the surrogate-side feature deviation
$\Delta \mathbf{z}_i := f_{\hat{\pi}}(\tilde{\mathbf{x}}_i)-f_{\hat{\pi}}(\mathbf{x}_i)$
and the target-side deviation
$\Delta \mathbf{g}_i := f_{\pi}(\tilde{\mathbf{x}}_i)-f_{\pi}(\mathbf{x}_i)$.

\begin{assumption}[Linear alignment with bounded residual]\label{asm:lin}
There exists a matrix $A^\star\!\in\!\mathbb{R}^{d\times d}$ such that
\begin{equation}
f_{\pi}(\mathbf{x}) \;=\; f_{\hat{\pi}}(\mathbf{x})\,A^\star \;+\; e(\mathbf{x}),
\end{equation}
where the alignment residual $e(\mathbf{x})$ satisfies
$\|\,e(\tilde{\mathbf{x}})-e(\mathbf{x})\,\|_2 \le \varepsilon_E$
for all pairs $(\mathbf{x},\tilde{\mathbf{x}})$ considered.
\end{assumption}

Assumption~\ref{asm:lin} states that the effect of a surrogate deviation must persist on the target with strength governed by $\sigma_{\min}(A^\star)$, the smallest singular value of the alignment map $A^\star$. The proposition below makes this dependence explicit.

\begin{proposition}[Lower-bounded target displacement]\label{prop:lb}
Under Assumption~\ref{asm:lin}, for any $(\mathbf{x}_i,\tilde{\mathbf{x}}_i)$
\begin{equation}
\label{eq:lb-l2-formal}
\big\|\Delta \mathbf{g}_i\big\|_2
\;\ge\;
\sigma_{\min}(A^\star)\,\big\|\Delta \mathbf{z}_i\big\|_2 \;-\; \varepsilon_E,
\end{equation}
and, using Hölder’s inequality $\|v\|_1 \le \sqrt{d}\|v\|_2$,
\begin{equation}
\label{eq:lb-l1-formal}
\big\|\Delta \mathbf{g}_i\big\|_1
\;\ge\;
\frac{\sigma_{\min}(A^\star)}{\sqrt{d}}\;\big\|\Delta \mathbf{z}_i\big\|_1 \;-\; \varepsilon_E.
\end{equation}
\end{proposition}


\noindent Proof is in \textit{Appendix}~\ref{sec:proof}. Proposition~\ref{prop:lb} links target-side deviation to the surrogate-side. Therefore, any strategy that enlarges $\|\Delta\mathbf{z}_i\|$ (\textit{e.g.,} via an $\ell_1$ objective) necessarily induces a nontrivial response on the target. Thus we can capture why we could use \textbf{L1 loss} $\mathcal{L}_1$ to maximize feature discrepancy:

\begin{corollary}[Effect of maximizing $\ell_1$ deviation]\label{cor:maxl1}
If an attack increases the surrogate-side $\ell_1$ deviation,
e.g.\ by maximizing $\mathcal{L}_{1}=\|\Delta \mathbf{z}_i\|_1$, then the target-side deviation obeys the linear lower bound in
Eq.~\ref{eq:lb-l1-formal}.
In particular, when the alignment is well-conditioned
($\sigma_{\min}(A^\star)$ not small) and the residual coupling $\varepsilon_E$ is modest, increasing $\|\Delta \mathbf{z}_i\|_1$ necessarily induces a nontrivial increase of
$\|\Delta \mathbf{g}_i\|_1$.
\end{corollary}



\noindent\textbf{Repulsive Contrastive Regularization.} Complementing the $\mathcal{L}_1$ deviation term, we introduce a
\emph{repulsive} contrastive objective that explicitly pushes the patched
feature $\tilde{\mathbf{z}}_i$ away from its clean anchor $\mathbf{z}_i$.
For each sample $i$, we still treat $(\mathbf{z}_i, \tilde{\mathbf{z}}_i)$ as
a distinguished pair and $\{\tilde{\mathbf{z}}_j\}_{j\neq i}$ as a reference set, and adopt the InfoNCE \citep{chen2020simple} as a repulsion term
\begin{equation}
\label{eq:con}
    \mathcal{L}_\mathrm{con}
    = -\,\frac{1}{N} \sum_{i=1}^{N}
    \log \frac{\exp(\mathrm{sim}(\mathbf{z}_i, \tilde{\mathbf{z}}_i)/\tau)}
    {\sum_{j=1}^{N} \exp(\mathrm{sim}(\mathbf{z}_i, \tilde{\mathbf{z}}_j)/\tau)},
\end{equation}
where $\mathrm{sim}$ denotes cosine similarity and $\tau$ is a temperature.
Maximizing (minimizing) $\mathcal{L}_\mathrm{con}$ encourages the similarity $\mathrm{sim}(\mathbf{z}_i, \tilde{\mathbf{z}}_i)$ to \emph{decrease (increase)}, effectively pushing $\tilde{\mathbf{z}}_i$ away (pulling $\tilde{\mathbf{z}}_i$ close) from its clean anchor and concentrating the change along directions that are consistently shared across the batch.

\vspace{1mm}
\noindent\textbf{Overall Feature-space Objective.}
Combining both components, we obtain the objective for $\Delta$ as given in Eq.~\ref{eq:feature_measurement}:
\begin{equation}
\label{eq:overall-inner}
    \mathcal{J}_\mathrm{tr}
    \;=\;
    \mathcal{L}_{1}
    \;+\;
    \lambda_\mathrm{con}\,\mathcal{L}_\mathrm{con},
\end{equation}
where $\mathcal{L}_{1}$ is the $\ell_1$ loss term and
$\mathcal{L}_\mathrm{con}$ is the repulsive contrastive objective, and
$\lambda>0$ balances their contributions.

\subsection{Robustness-augmented Universal Patch Attack}
\label{sec:dual}
\noindent\textbf{Emulate Robust Surrogates without Retraining VLAs.}
Transfer-based attacks on image classifiers have shown that adversarial examples generated on \emph{adversarially trained} or \emph{slightly robust} source models transfer significantly better than those crafted on standard models~\citep{springer2021little,jones2022if}.
Robust training encourages the source model to rely on more ``universal'' features shared across architectures, so perturbations aligned with these features exhibit stronger cross-model transferability.
A natural strategy would be to use an adversarially trained VLA as the surrogate.
However, adversarially trained large VLA policies is practically prohibitive: it requires massive interactive data and computing, and can substantially degrade task performance.
Instead, we still optimize a \emph{single universal physical patch}, but
augment it with a \emph{sample-wise, invisible} perturbation that \emph{emulates}
adversarial (robust) training on the surrogate.
This perturbation is applied globally and updated to \emph{counteract} patch-induced feature deviations, effectively ``hardening'' the surrogate along the directions the patch tries to exploit.
Since the universal patch is localized while the sample-wise perturbations remain invisible and input-specific, their interference is limited, and the patch can then exploit the robust feature directions revealed by this
hardening step.

\vspace{1mm}
\noindent\textbf{Bi-level Robustness-augmented Optimization.}
Formally, let $\boldsymbol{\delta}$ denote the universal patch and $\boldsymbol{\sigma}$ a sample-wise perturbation confined to the patch mask.
Given a optimizing loss $\mathcal{J}_{\mathrm{tr}}$ on the surrogate
$\hat{\pi}$, we consider the following
robustness-augmented bi-level objective:
\begin{equation}
\label{eq:bilevel-rapa}
\!\!\!\begin{aligned}
&\boldsymbol{\delta}^{\star}
\in
\arg\max_{\mathcal{S}(\boldsymbol{\delta}) < \rho}
  \mathbb{E}_{\substack{\mathbf{x}\sim p(\mathbf{x})\\T_t\sim\mathcal{T}}}
    \mathcal{J}_{\mathrm{tr}}\!\Big(
        \mathcal{P}(\mathbf{x}\!+\!\boldsymbol{\sigma}^{\star}(\mathbf{x}),\boldsymbol{\delta},T_t);\hat{\pi}
    \Big)
\\
&\text{s.t.}~
\boldsymbol{\sigma}^{\star}(\mathbf{x})
\in
\arg\!\!\!\!\min_{\|\boldsymbol{\sigma}\|_{\infty}\le\epsilon_{\sigma}}
\mathbb{E}_{\substack{\mathbf{x}\sim p(\mathbf{x})\\T_t\sim\mathcal{T}}}
    \mathcal{J}_{\mathrm{tr}}\!\Big(\mathcal{P}(
        \mathbf{x}\!+\!\boldsymbol{\sigma},\boldsymbol{\delta},T_t);\hat{\pi}
    \Big).
\end{aligned}
\end{equation}
The inner problem ``adversarially trains'' the surrogate locally by finding a small, sample-wise perturbation to \emph{reduce} the attack loss, and the outer problem then maximizes the same loss with respect to $\boldsymbol{\delta}$ in this hardened neighborhood. 

To further strengthen transferability, the outer maximization is not driven by the feature displacement alone.
In the following subsections, we introduce \textbf{additional loss components} that shape \emph{where} the model attends and \emph{what} semantics the patch encodes, and jointly optimize them within this robustness-augmented framework.

\subsection{Patch Attention Dominance: Cross-Modal Hijack Loss}
\label{sec:attn}
\noindent\textbf{Action-relevant Queries as the Attack Handle.}
In VLA policies, actions are largely driven by a small set of
\emph{action-relevant} text queries whose cross-modal attention to vision
decides which visual regions control the policy.
Our universal patch is therefore designed as a \emph{location-agnostic
attention attractor}: regardless of placement, skew, or orientation, it should
\textbf{redirect the attention} of these action-relevant queries \emph{from true semantic regions to the patch}.
Concretely, we aim to \emph{increase} the attention increments on the patch
vision tokens while \emph{reducing} increments on non-patch tokens, based on the difference between patched and clean runs under random placements.

\vspace{1mm}
\noindent\textbf{Patch-induced Attention Increments for Action-relevant queries.}
From clean and patched runs, we collect the last $N$ attention blocks $\mathbf{A}$ from $f_\mathrm{llm}$, average
over heads, and slice out the text$\to$vision submatrix via $\mathrm{tv}(\cdot)$:
\begin{equation}
\begin{aligned}
\bar{\mathbf{A}}^{(l)}_c &= \tfrac{1}{H}\!\sum_{h=1}^{H}\mathbf{A}^{(l)}_{c,:,h,:,:},\quad
\mathbf{B}^{(l)}_c=\mathrm{tv}\!\big(\bar{\mathbf{A}}^{(l)}_c\big),\\
\bar{\mathbf{A}}^{(l)}_p &= \tfrac{1}{H}\!\sum_{h=1}^{H}\mathbf{A}^{(l)}_{p,:,h,:,:},\quad
\mathbf{B}^{(l)}_p=\mathrm{tv}\!\big(\bar{\mathbf{A}}^{(l)}_p\big),
\end{aligned}
\end{equation}
where $l=L-N+1,\dots,L$ indexes the last $N$ layers.
We row-normalize over vision tokens (index $p$) and average across layers to
obtain attention \emph{shares}, then define the patch-induced share increment:
\begin{equation}
\boldsymbol{\Delta}
= \tfrac{1}{N}\!\sum_{l}\mathrm{rn}\!\big(\mathbf{B}^{(l)}_p\big)
 -\tfrac{1}{N}\!\sum_{l}\mathrm{rn}\!\big(\mathbf{B}^{(l)}_c\big)
\in\mathbb{R}^{B\times T\times P},
\end{equation}
where $\mathrm{rn}(\cdot)$ denotes row-normalization over $p$.
By optimizing $\boldsymbol{\Delta}$ rather than raw attention, the objective
depends only on \emph{patch-induced} changes.

\vspace{1mm}
\noindent\textbf{Action-relevant Queries.} To focus precisely on action-relevant queries and avoid surrogate-specific
overfitting, we restrict the optimization to the top-$\rho$ text tokens (per batch) that already receive the highest clean attention:
\begin{equation}
\tilde{\boldsymbol{\Delta}}=\boldsymbol{\Delta}\odot\boldsymbol{\chi},\qquad
\boldsymbol{\chi}=\mathrm{TopKMask}\!\big(\mathbf{B}_c;\,\rho\big),
\end{equation}
where $\mathrm{TopKMask}$ returns a binary mask over the text positions,
broadcast across vision tokens.
These top-$\rho$ tokens are our proxy for action-relevant queries.

\vspace{1mm}
\noindent\textbf{Patch vs.\ Non-patch Attention Increments.}
To capture the effect of patch location on visual tokens, we map the
pixel-level mask $\mathbf{M}_{T_t}\in\{0,1\}^{H\times W}$ to a token-level
mask $\mathbf{M}_z\in[0,1]^p$ via bilinear interpolation, where $p$ is the
number of visual tokens (\textit{e.g.,} $p=g^2$ for a $g\times g$ ViT grid), and then
flatten it to length $p$.
We then aggregate the attention increments routed from action-relevant
queries into patch versus non-patch vision tokens:
\begin{equation}
\begin{aligned}
d_{\mathrm{patch}}&=\langle \tilde{\boldsymbol{\Delta}},\,\mathbf{M}_z\rangle_{p},
~d_{\mathrm{non}}  =\langle \tilde{\boldsymbol{\Delta}},\,\mathbf{1}-\mathbf{M}_z\rangle_{p},\\
\mathrm{non\_top} &=\max\nolimits_{p}\big(\tilde{\boldsymbol{\Delta}}\odot(\mathbf{1}-\mathbf{M}_z)\big),
\end{aligned}
\end{equation}
where $\langle\cdot,\cdot\rangle_p$ sums over the vision index $p$.
$d_{\mathrm{patch}}$ ($d_{\mathrm{non}}$) measures how much extra
attention the patch induces on patch (non-patch) tokens from
action-relevant text queries.

\vspace{1mm}
\noindent\textbf{Patch Attention Dominance (PAD) Loss.}
Finally, we define the attention-hijack objective to maximize by explicitly
\emph{increasing} patch-related increments and \emph{decreasing} non-patch
increments, with a margin against the strongest non-patch route:
\begin{equation}
\begin{aligned}
\mathcal{L}_{\mathrm{PAD}}
= &\,\mathbb{E}[d_{\mathrm{patch}}]
- \lambda\,\mathbb{E}\big[\mathrm{ReLU}(d_{\mathrm{non}})\big] \\
&- \mathbb{E}\!\Big[\mathrm{ReLU}\!\big(m-(d_{\mathrm{patch}}-\mathrm{non\_top})\big)\Big],
\end{aligned}
\label{eq:padloss}
\end{equation}
where $\mathbb{E}[\cdot]$ averages over the selected (action-relevant) text
tokens.
The first term increases patch attention increments, the second penalizes
positive increments on non-patch tokens, and the margin term
enforces that the patch’s increment exceeds the strongest non-patch increment
by at least $m$.
Together, these terms induce \emph{Patch Attention Dominance}, where
action-relevant queries direct their additional attention to the patch rather
than true semantic regions.

\subsection{Patch Semantic Misalignment: Text-Similarity Attack Loss}
\label{sec:text}

\noindent\textbf{Semantic Steering beyond Attention.}
Merely hijacking cross-modal attention does not guarantee a consistent
behavioral bias across models or tasks.
To further enhance transferability, we constrain the patch also in
\emph{semantic} space: we \textbf{steer} the visual representation of patch-covered
tokens \textbf{toward} a set of cross-model-stable action/direction primitives
(\emph{probe phrases}), while simultaneously\textbf{ pushing it away from the holistic representation} of the current instruction.
The probes (\textit{e.g.,} ``put'', ``pick up'', ``place'', ``open'', ``close'', ``left'', ``right'') act as architecture-agnostic anchors, and the repulsion
from the instruction embedding induces a persistent, context-dependent
semantic misalignment that more reliably derails the policy decoder.

\vspace{1mm}
\noindent\textbf{Patch Pooling and Semantic Anchors.}
Let $\mathbf z_j\in\mathbb R^D$ be visual token features and
$m_j\!\in\!\mathbf{M}_z$ the corresponding patch-token mask.
We pool and $\ell_2$–normalize the patch feature:
\begin{equation}
\label{eq:ts-pool}
\hat{\mathbf v}_{\mathrm{patch}}
= \Bigg\|
      {\bigg(\sum_{j=1}^{P} m_j\,\mathbf z_j\bigg)}\Big/{\bigg(\sum_{j=1}^{P} m_j+\varepsilon\bigg)}
  \Bigg\|_2.
\end{equation}
Let $\{\hat{\mathbf p}_k\}_{k=1}^{K}$ be normalized \emph{probe prototypes}
(\textit{e.g.,} action and direction anchors), and let $\hat{\mathbf t}$ denote a
normalized representation of the whole current instruction (\textit{e.g.,} the mean of
the last-layer text states from $f_\mathrm{llm}$).

\vspace{1mm}
\noindent\textbf{Patch Semantic Misalignment (PSM) Loss.} We then define the text-similarity attack loss to maximize as
\begin{equation}
\label{eq:ts-loss}
\mathcal L_{\text{PSM}}
= \alpha\!\left[
    \log\!\sum_{k=1}^{K}
    \exp\!\Big(\frac{\hat{\mathbf v}_{\mathrm{patch}}^{\!\top}\hat{\mathbf p}_k}{\tau}\Big)
  \right]
\;-\; \beta\,\hat{\mathbf v}_{\mathrm{patch}}^{\!\top}\hat{\mathbf t},
\end{equation}
with temperatures $\tau>0$ and weights $\alpha,\beta>0$.

Eq.~\ref{eq:ts-pool} yields a location-agnostic semantic descriptor for the
patch-covered tokens.
In Eq.~\ref{eq:ts-loss}, the first (LogSumExp) term \emph{pulls}
$\hat{\mathbf v}_{\mathrm{patch}}$ toward any probe prototype, avoiding
dependence on a single phrase while focusing gradients on the most compatible
anchors as $\tau$ decreases.
The second term \emph{pushes} the patch feature away from the instruction
embedding, inducing a persistent, context-dependent semantic mismatch, with
$\alpha,\beta$ balancing pull and push.
The loss is fully differentiable w.r.t.\ patch parameters via $\mathbf z_j$ and
complements attention hijacking by steering the \emph{attended} content toward
a stable, transferable semantic direction.

\subsection{Universal Patch Attack via Robust Feature, Attention, and Semantics (UPA-RFAS)}
\label{sec:alg}
The overall optimization process is in Algorithm~\ref{alg:raupa} where:

\vspace{1mm}
\noindent\textbf{Inner Minimization.}
Given $\mathbf{x}$ at time $t$ and the current patch $\boldsymbol{\delta}$,
we initialize a global invisible perturbation $\boldsymbol{\sigma}^{(0)} = \mathbf{0}$
and update it by Projected Gradient Descent (PGD)~\cite{madry2017towards}:
\begin{equation}
\boldsymbol{\sigma}^{(i+1)}
  \!\leftarrow\!\!
  \mathop{\Pi}_{\|\cdot\|_{\infty}\le\epsilon_{\sigma}}
  \!\!\!\bigg(\!\!
    \boldsymbol{\sigma}^{(i)}
    \!-\! \eta_{\boldsymbol{\sigma}}
      \nabla_{\boldsymbol{\sigma}}
      \mathcal{J}_{\mathrm{in}}\Big(\!\mathcal{P}(
        \mathbf{x}\!+\!\boldsymbol{\sigma}^{(i)},\boldsymbol{\delta},T_t);\hat{\pi}
    \!\Big)
  \!\!\bigg),
\end{equation}
where $\mathcal{J}_{in}=\mathcal{J}_{tr}$ is in Eq.~\ref{eq:overall-inner}, $\Pi_{\|\cdot\|_{\infty}\le\epsilon_{\sigma}}$ projects onto the
$\ell_{\infty}$ ball of radius $\epsilon_\sigma$,  $\eta_{\boldsymbol{\sigma}}$ is the step size, and $T_t$ is sampled once from $\mathcal{T}$ across iterations. $\boldsymbol{\sigma}^{\star}$ is the final perturbation.

\vspace{1mm}
\noindent\textbf{Outer Maximization.}
With $\boldsymbol{\sigma}^{\star}(\mathbf{x})$ fixed, we update the universal patch
$\boldsymbol{\delta}$ by AdamW~\cite{loshchilov2017decoupled} to maximize the objective with additional losses under
randomized transformations:
\begin{equation}
\begin{split}
  &\!\boldsymbol{\delta}
  \!\leftarrow\!
  \mathrm{AdamW}\bigg(\!\!\!-\!\mathcal{J}_{\mathrm{out}}\Big(\mathcal{P}(
        \mathbf{x}\!+\!\boldsymbol{\sigma}^{\star}(\mathbf{x}),\boldsymbol{\delta},T_t);\hat{\pi}
    \Big);\eta_{\boldsymbol{\delta}}
  \!\bigg), \\
  &\!\mathcal{J}_\mathrm{out}
    \!=\!
    \mathcal{L}_{1}
    \!+\!
    \lambda_\mathrm{con}\,\mathcal{L}_\mathrm{con}\!+\!
    \lambda_\mathrm{PAD}\,\mathcal{L}_\mathrm{PAD}\!+\!
    \lambda_\mathrm{PSM}\,\mathcal{L}_\mathrm{PSM},
    \label{eq:overall-outer}
\end{split}
\end{equation}
where the patch $\boldsymbol{\delta} \in [0,1]^{h_p \times w_p \times 3}$ respects the area budget, and $\eta_{\boldsymbol{\delta}}$ is the learning rate. At each iteration, we sample $T_t \sim \mathcal{T}$ and clamp $\boldsymbol{\delta}$ to the valid range $[0,1]$.

\begin{algorithm}[t]
\caption{UPA-RFAS}
\label{alg:raupa}
\begin{algorithmic}[1]
\STATE \textbf{Input:} surrogate $f_{\hat{\pi}}$, subset $\mathcal{D}_s$, universal patch $\boldsymbol{\delta} \in [0,1]^{h_p \times w_p \times 3}$,
budget $\epsilon_\sigma$, inner steps $I$, outer steps $K$,
step sizes $\eta_{\boldsymbol{\sigma}},\eta_{\boldsymbol{\delta}}$, weights $\lambda_{\mathrm{con}},\lambda_{\mathrm{PAD}},\lambda_{\mathrm{PSM}}$
\FOR{mini-batch data $(\mathbf{x},c,t) \subset \mathcal{D}_s$}
\STATE \textcolor{teal}{\# Inner minimization}
\STATE Initialize sample-wise perturbation $\boldsymbol{\sigma}^{(1)} \leftarrow \mathbf{0}$ \\
\STATE Sample $T_t\sim\mathcal{T}$\\
\FOR{$i = 1$ \TO $I$}
\STATE $\mathcal{J}_{\mathrm{in}}\!=\!\mathcal{J}_{\mathrm{tr}}\Big(\mathcal{P}(
        \mathbf{x}+\boldsymbol{\sigma}^{(i)},\boldsymbol{\delta},T_t);\hat{\pi}
    \Big)$ via Eq.~\ref{eq:applypatch-compact} and \ref{eq:overall-inner}
\STATE $\boldsymbol{\sigma}^{(i+1)} \leftarrow
\Pi_{\|\cdot\|_\infty \le \epsilon_\sigma}
\big(\boldsymbol{\sigma}^{(i)}
- \eta_\sigma \nabla_{\boldsymbol{\sigma}} \mathcal{J}_{\mathrm{in}}\big)$
\ENDFOR \\
\STATE $\boldsymbol{\sigma}^{\star} \leftarrow \boldsymbol{\sigma}^{(I)}$\\
\vspace{0.5mm}
\textcolor{teal}{\#Outer maximization}
\FOR{$k = 1$ \TO $K$}
\STATE Sample $T_t\sim\mathcal{T}$\\
\STATE Compute $\mathcal{J}_\mathrm{out}$ via  Eq.~\ref{eq:applypatch-compact}, \ref{eq:padloss}, \ref{eq:ts-loss} and \ref{eq:overall-outer}
\STATE $\boldsymbol{\delta}
  \!\leftarrow\!
  \mathrm{AdamW}\bigg(\!\!-\!\mathcal{J}_{\mathrm{out}}\Big(\!\mathcal{P}(
        \mathbf{x}\!+\!\boldsymbol{\sigma}^{\star}(\mathbf{x}),\boldsymbol{\delta},T_t);\hat{\pi}
    \!\Big);\eta_{\boldsymbol{\delta}}
  \!\bigg)$
  \STATE $\boldsymbol{\delta}
  \leftarrow\mathrm{Clip}_{[0,1]}(\boldsymbol{\delta})$
\ENDFOR
\ENDFOR
\STATE \textbf{return} $\boldsymbol{\delta}$
\end{algorithmic}
\end{algorithm}

\section{Experiments}

\begin{table*}[t]
\centering
\small
\caption{Task success rate (\%) when transferring from the surrogate OpenVLA-7B to different victim models on LIBERO.}
\setlength{\tabcolsep}{1.5pt}
\vspace{-3mm}
\begin{tabular}{c||ccccc|ccccc||ccccc|ccccc}
\toprule
\multirow{3}{*}{objective}
& \multicolumn{10}{c||}{\textbf{Victim: OpenVLA-oft-w}}
& \multicolumn{10}{c}{\textbf{Victim: OpenVLA-oft}} \\
\cmidrule(lr){2-11}\cmidrule(lr){12-21}
& \multicolumn{5}{c|}{Simulated} & \multicolumn{5}{c||}{Physical}
& \multicolumn{5}{c|}{Simulated} & \multicolumn{5}{c}{Physical} \\
\cmidrule(lr){2-6}\cmidrule(lr){7-11}\cmidrule(lr){12-16}\cmidrule(lr){17-21}
& spatial & object & goal & long & avg.
& spatial & object & goal & long & avg.
& spatial & object & goal & long & avg.
& spatial & object & goal & long & avg. \\
\midrule
Benign
& 99 & 99 & 98 & 97 & 98.25
& 99 & 99 & 98 & 97 & 98.25
& 98 & 98 & 98 & 94 & 97.00
& 98 & 98 & 98 & 94 & 97.00 \\

$\text{UMA}_1$
& 25 & 86 & 40 & 31 & 45.50
& 83 & 89 & 76 & 73 & 80.25
& 79 & 95 & 69 & 3 & 61.50
& 96 & 90 & 90 & 83 & 89.75 \\

$\text{UMA}_{1-3}$
& 46 & 88 & 38 & 39 & 52.75
& 90 & 87 & 83 & 81 & 85.25
& 90 & 93 & 57 & 3 & 60.75
& 96 & 81 & 93 & 80 & 87.50 \\

$\text{UADA}_{1}$
& 35 & 82 & 27 & 21 & 41.25
& 71 & 90 & 57 & 74 & 73.00
& 94 & 86 & 61 & 3 & 61.00
& 92 & 96 & 79 & 84 & 87.75 \\

$\text{UADA}_{1-3}$
& 37 & 74 & 21 & 33 & 41.25
& 65 & 88 & 46 & 61 & 65.00
& 87 & 90 & 64 & 4 & 61.25
& 92 & 95 & 61 & 90 & 84.50 \\

$\text{TMA}_1$
& 69 & 89 & 58 & 61 & 69.25
& 78 & 92 & 74 & 83 & 81.75
& 99 & 93 & 81 & 25 & 74.50
& 98 & 92 & 84 & 86 & 90.00 \\

$\text{TMA}_7$
& 47 & 78 & 47 & 34 & 51.50
& 90 & 96 & 89 & 90 & 91.25
& 83 & 93 & 75 & 62 & 78.25
& 97 & 88 & 89 & 87 & 90.25 \\
\midrule
Our
& \textbf{7} & \textbf{0} & \textbf{10} & \textbf{6} & \textbf{5.75}
& \textbf{26} & \textbf{53} & \textbf{54} & \textbf{28} & \textbf{40.25}
& \textbf{66} & \textbf{43} & \textbf{62} & \textbf{3} & \textbf{43.50}
& \textbf{69} & \textbf{74} & \textbf{76} & \textbf{27} & \textbf{61.50} \\
\bottomrule
\end{tabular}
\label{tab:libero_victim_compare}
\vspace{-2mm}
\end{table*}

\noindent\textbf{Datasets. }
We evaluate our attacks on BridgeData V2 \citep{walke2023bridgedata} and LIBERO \citep{liu2023libero} using the corresponding VLA models. BridgeData V2 is a real-world corpus spanning 24 environments and 13 manipulation skills (\textit{e.g.,} grasping, placing, object rearrangement), comprising 60,096 trajectories. LIBERO is a simulation suite with four task families-Spatial, Object, Goal, and Long, where LIBERO-Long combines diverse objects, layouts, and extended horizons, making multi-step planning particularly challenging. 

\vspace{1mm}
\noindent\textbf{Baseline. }
We adopt RoboticAttack \citep{wang2025exploring}’s 6 objectives as the baselines, including Untargeted Manipulation Attack (UMA), Untargeted Action Discrepancy Attack (UADA), and Targeted Manipulation Attack (TMA) corresponding to different Degree-of-freedom (DoF). For each, experiments follow the original loss definitions and evaluation protocol. We further consider both \textbf{simulated and physical victim settings}: a model trained in simulation on the LIBERO-Long suite using the \emph{OpenVLA-7B-LIBERO-Long} variant, and a model trained on real-world BridgeData v2 data with the \emph{OpenVLA-7B} model, respectively.

\vspace{1mm}
\noindent\textbf{Surrogate and Victim VLAs. }
We evaluate universal, transferable patches under a strict black-box transfer protocol. Surrogate models are chosen from publicly available, widely used VLA \citep{kim2024openvla} to reflect prevailing design trends. The primary surrogate models are \emph{OpenVLA-7B} trained on physical dataset BridgeData V2 \citep{walke2023bridgedata} and \emph{OpenVLA-7B-LIBERO-Long} fine-tuned on LIBERO-Long. During transfer, \textbf{no information about victim models} is used, including weights, architecture details beyond public model names, fine-tuning datasets, recipes, or hyperparameters. Specifically, we select \emph{OpenVLA-oft} \citep{kim2025fine} and $\pi$ series \citep{black2024pi_0, black2024pi05} models as victim models. 
Built on OpenVLA, \emph{OpenVLA-oft} introduces an optimized fine-tuning recipe that notably improves success rates (from 76.5\% to 97.1\%) and delivers $\sim$26$\times$ throughput. To stress cross-recipe and cross-task vulnerability, we test on four variants fine-tuned on four distinct LIBERO task suites, as well as a multi-suite model trained jointly on all four (\emph{OpenVLA-oft-w}).
The \emph{$\pi$} family differs fundamentally from OpenVLA in backbone choice, pretraining/fine-tuning data, and training strategy, making transfer substantially harder. We therefore assess black-box transfer on $\pi_0$ \citep{black2024pi_0}, which provide a stringent test of model-agnostic patch behavior across heterogeneous VLA designs. 

\vspace{1mm}
\noindent\textbf{Implementation \& Evaluation Details. }
We evaluate on the LIBERO benchmark \citep{liu2023libero}. Each suite contains 10 tasks, and each task is attempted in 10 independent trials, yielding 100 rollouts per suite, following \citep{black2024pi_0}. Consistent with \citep{wang2025exploring}, patch placement sites are predetermined for each suite to avoid occluding objects in the test scenes. More implementation details can be found in the \textit{Appendix}~\ref{sec:im}. Regarding the evaluation metric, we adopt the concept of Success Rate (SR) introduced in LIBERO \citep{liu2023libero} across all setting.

\subsection{Main Results}
We first evaluate the white-box performance of our patches, where the victim model is identical to the surrogate. The results in \textit{Appendix}~\ref{sec:white} demonstrate that our method achieves strong white-box attack capability.
For the \emph{OpenVLA-7B} \citep{kim2024openvla} to \emph{OpenVLA-oft-w} \citep{kim2025fine} transfer experiment, Tab.~\ref{tab:libero_victim_compare} shows that our patch objective induces the strongest degradation of task success rates. In the simulated setting, the clean policy succeeds on 98.25\% of tasks on average, while our method reduces the success rate to only 5.75\%, corresponding to more than a 92\% point drop. Existing objectives such as UMA, UADA, and TMA do transfer to the victim but remain much less destructive: their average success rates stay between 41.25\% and 69.25\%, and they leave certain categories almost intact, for example, object-centric tasks still above 74\% success for UMA and UADA. In contrast, our patch almost completely disables the policy across all four task types. Tab.~\ref{tab:libero_victim_compare} also reports the attack results under physical setting. A similar trend appears: all baselines still retain high average success (65.00\%-91.25\%), whereas our method again yields the lowest success rate of 40.25\%. This indicates that our patch objective not only transfers more effectively to the simulated environments, but also produces substantially stronger degradation under the physical environment, establishing a consistently harder universal patch baseline across both settings.

Beyond the transfer from \emph{OpenVLA-7B} to \emph{OpenVLA-oft-w}, we further evaluate transfer to four different \emph{OpenVLA-oft} variants that are separately fine-tuned on different LIBERO task suites, creating a larger distribution and policy gap from the surrogate. Tab.~\ref{tab:libero_victim_compare} shows that our objective still achieves consistently stronger transfer than all baselines across both simulated and physical setups, highlighting the effectiveness of our design. \textbf{Additional transfer results}, including attacks transferred to $\pi_0$, are in \textit{Appendix}~\ref{sec:pi0}, and show that our methods still enhance attacks in the most challenging case of transferring to entirely different VLAs.

\subsection{Ablation Study}

\begin{table}[]
    \centering
    \caption{Ablation for transfer to openvla-oft under physical setting.}
    \vspace{-3mm}
\begin{tabular}{c||ccccc}
\toprule
objective & spatial & object & goal & long & avg. \\ \midrule
Our &69 &74 & 76 & 27 & 61.50 \\ 
w/o RUPA & 70 & 75 & 71 & 33 & 62.25 \\
w/o PAD & 68 & 67 & 77& 38 &62.50\\
w/o PSM &69 & 72 & 81 & 32 & 63.50 \\
w/o $\mathcal{J}_\mathrm{tr}$ &90 &86 & 94&73& 85.75\\
w/o $\mathcal{L}_\mathrm{con}$ & 93 & 63 & 79 &48 &70.75  \\
w/o $\mathcal{L}_1$ &74&74&77&31& 64.00\\
\bottomrule
\end{tabular}
\label{tab:ab}
\vspace{-2mm}
\end{table}

\begin{table}[]
    \centering
    \caption{Ablation on text-probe phrasing for transfer to openvla-oft in the physical setting.}
    \vspace{-3mm}
\begin{tabular}{c|ccccc}
\toprule
objective &  spatial & object & goal & long & avg. \\ \midrule
Our & 69 &74 & 76 & 27 & 61.50\\
Action &76& 67&94 &48 &71.25 \\
Direction & 72 &75& 78&75 &75.00 \\
\bottomrule
\end{tabular}
\label{tab:text}
\vspace{-2mm}
\end{table}

\noindent\textbf{Impact of Each Design.}
Tab.~\ref{tab:ab} further validates the role of each component in our objective.
Dropping any single module (RUPA, PAD, or PSM) consistently weakens the attack, reflected by higher average success rates than the full model. The most severe degradation appears in the \emph{w/o}~$\mathcal{J}\mathrm{tr}$ variant, where the average success rate jumps to 85.75\%, close to the benign and baseline levels. Since $\mathcal{J}\mathrm{tr}$ jointly contains both $\mathcal{L}1$ and $\mathcal{L}{\mathrm{con}}$ (\textit{i.e.,} it removes the entire first-stage feature-space minimization), this indicates that our feature-space $\ell_1$ and contrastive misalignment objectives, together with the RUPA designs, are essential for strong transfer.
Moreover, the impact of $\mathcal{L}_\mathrm{con}$ is noticeably larger than that of $\mathcal{L}_1$. By Prop.~\ref{prop:lb} and Cor.~\ref{cor:maxl1}, $\mathcal{L}1$ is a distance-based term that mainly controls the \emph{magnitude} of the surrogate deviation, whereas $\mathcal{L}_\mathrm{con}$, built on cosine similarity, focuses on feature angles and thus shapes the \emph{direction} of the displacement. Consequently, even without $\mathcal{L}1$, $\mathcal{L}\mathrm{con}$ can still drive patched features away from their clean anchors along transferable directions, so the attack remains relatively strong.

\vspace{1mm}
\noindent\textbf{Impact of Text Probes.}
Tab.~\ref{tab:text} analyzes how text-probe phrasing influences transfer in the physical setting.
We compare our default probes, which jointly encode both action and spatial direction, against two reduced variants: \textbf{Action} probes that only include verbs (\textit{e.g.,} “put”, “pick up”, “place”, “turn on”, “push”, “open”, “close”) and \textbf{Direction} probes that only contain spatial words (\textit{e.g.,} “left”, “right”, “bottom”, “back”, “middle”, “top”, “front”).
Using action-only or direction-only probes markedly weakens the attack: the average success rate increases to 71.25\% and 75.00\%, respectively, compared to 61.5\% with our design. This suggests that jointly encoding action and directional cues produces text queries that more closely match the policy’s action-relevant channels, thereby enabling more effective cross-model transfer.
Ablation study of more specific parameters can be found in \textit{Appendix}~\ref{sec:dab}.

\subsection{Patch Pattern Analysis}
\begin{figure}[t]
  \centering
  \begin{subfigure}[b]{0.32\linewidth}
    \centering
    \includegraphics[width=\linewidth]{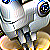}
    \subcaption{UADA${_{1\text{-}3}}$} 
    \label{fig:patch:a}
  \end{subfigure}\hfill
  \begin{subfigure}[b]{0.32\linewidth}
    \centering
    \includegraphics[width=\linewidth]{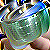}
    \subcaption{TMA with DoF 7} 
    \label{fig:patch:c}
  \end{subfigure}\hfill
\begin{subfigure}[b]{0.32\linewidth}
    \centering
    \includegraphics[width=\linewidth]{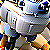}
    \subcaption{\textbf{Ours}} 
    \label{fig:patch:e}
  \end{subfigure}\hfill
  \begin{subfigure}[b]{0.32\linewidth}
    \centering
    \includegraphics[width=\linewidth]{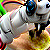}
    \subcaption{UADA${_{1\text{-}3}}$} 
    \label{fig:patch:b}
  \end{subfigure}\hfill
  \begin{subfigure}[b]{0.32\linewidth}
    \centering
    \includegraphics[width=\linewidth]{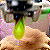}
    \subcaption{TMA with DoF 7} 
    \label{fig:patch:d}
  \end{subfigure}\hfill
  \begin{subfigure}[b]{0.32\linewidth}
    \centering
    \includegraphics[width=\linewidth]{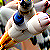}
    \subcaption{\textbf{Ours}} 
    \label{fig:patch:f}
  \end{subfigure}
\vspace{-2mm}
  \caption{\textbf{Patch visualization and comparison.} The first row is trained in a simulated setting, and the second row is trained in a physical setting.}
  \vspace{-2mm}
  \label{fig:patch}
\end{figure}

As shown in Fig.~\ref{fig:patch}, we can see that baseline end-to-end methods \citep{wang2025exploring} produce scene-tied patterns: UADA yields textures that closely resemble the robot gripper in both simulation and physical settings (Fig.~\ref{fig:patch:a} and~\ref{fig:patch:b}), while TMA generates more abstract yet surrogate-specific shapes (Fig.~\ref{fig:patch:c} and~\ref{fig:patch:d}). These behaviors indicate overfitting to object/embodiment cues, which hampers cross-model and cross-setting transfer. In contrast, our universal transferable patch (Fig.~\ref{fig:patch:e} and~\ref{fig:patch:f}) is learned in feature space to perturb higher-level, model-agnostic representations shared across VLAs. By jointly optimizing feature-space, attention, and semantic objectives, our patch combines the strengths of prior designs, avoids object mimicry, and yields a universal patch that reliably transfers across tasks, embodiments, and environments, resulting in stronger black-box transfer.

\section{Conclusion}
In this paper, we present the first study of universal, transferable patch attacks on VLA-driven robots and introduce UPA-RFAS, a unified framework that couples an $\ell_1$ feature deviation with repulsive contrastive alignment to steer perturbations into model-agnostic, high-transfer directions. UPA-RFAS integrates a robustness-augmented patch optimization and two VLA-specific losses, Patch Attention Dominance and Patch Semantic Misalignment, which achieve strong black-box transfer across models, tasks, and sim-to-real settings, revealing a practical patch-based threat and a solid baseline for future defenses.

\section*{Acknowledgement}
This work was carried out at the Rapid-Rich Object Search (ROSE) Lab, School of Electrical \& Electronic Engineering, Nanyang Technological University (NTU), Singapore. This research is supported by the National Research Foundation, Singapore and Infocomm Media Development Authority under its Trust Tech Funding Initiative and the DSO National Laboratories, Singapore, under the project agreement No. DSOCL25023. Any opinions, findings and conclusions or recommendations expressed in this material are those of the author(s) and do not reflect the views of National Research Foundation, Singapore and Infocomm Media Development Authority.

{
    \small
    \bibliographystyle{ieeenat_fullname}
    \bibliography{main}
}
\clearpage
\setcounter{page}{1}
\maketitlesupplementary

\setcounter{section}{0}
\renewcommand{\thesection}{\Alph{section}}

\section{Proof for Proposition 1}
\label{sec:proof}

\noindent\textbf{Proof.}
By Assumption~1, there exists a matrix $A^\star\in\mathbb{R}^{d\times d}$ and a
residual term $e(\cdot)$ such that, for every $\mathbf{x}$,
\begin{equation}
\label{eq:alignment_assumption}
f_{\pi}(\mathbf{x}) = f_{\hat{\pi}}(\mathbf{x})A^\star + e(\mathbf{x}),
\end{equation}
and for all pairs $(\mathbf{x},\tilde{\mathbf{x}})$ under consideration the residual difference is uniformly bounded:
\begin{equation}
\label{eq:residual_bound}
\bigl\| e(\tilde{\mathbf{x}})-e(\mathbf{x}) \bigr\|_2 \;\leq\; \varepsilon_E .
\end{equation}

\paragraph{Step 1: Expressing the target deviation.}
For a fixed pair $(\mathbf{x}_i,\tilde{\mathbf{x}}_i)$, denote the residual
difference by
\[
\Delta\mathbf{e}_i := e(\tilde{\mathbf{x}}_i)-e(\mathbf{x}_i).
\]
Using~\eqref{eq:alignment_assumption},
\begin{align*}
\Delta\mathbf{g}_i
&= f_{\pi}(\tilde{\mathbf{x}}_i) - f_{\pi}(\mathbf{x}_i) \\
&= \bigl(f_{\hat{\pi}}(\tilde{\mathbf{x}}_i)A^\star + e(\tilde{\mathbf{x}}_i)\bigr)
   - \bigl(f_{\hat{\pi}}(\mathbf{x}_i)A^\star + e(\mathbf{x}_i)\bigr) \\
&= \bigl(f_{\hat{\pi}}(\tilde{\mathbf{x}}_i)-f_{\hat{\pi}}(\mathbf{x}_i)\bigr)A^\star
   + \bigl(e(\tilde{\mathbf{x}}_i)-e(\mathbf{x}_i)\bigr) \\
&= \Delta\mathbf{z}_i A^\star + \Delta\mathbf{e}_i .
\end{align*}

\paragraph{Step 2: Lower-bounding the $\ell_2$ norm.}
Applying the reverse triangle inequality to $\Delta\mathbf{g}_i$ gives
\begin{equation}
\label{eq:triangle_step}
\|\Delta\mathbf{g}_i\|_2
= \|\Delta\mathbf{z}_i A^\star + \Delta\mathbf{e}_i\|_2
\;\geq\;
\|\Delta\mathbf{z}_i A^\star\|_2 - \|\Delta\mathbf{e}_i\|_2 .
\end{equation}
By the residual bound~\eqref{eq:residual_bound}, we have
$\|\Delta\mathbf{e}_i\|_2 \leq \varepsilon_E$.

Next, recall the standard singular value inequality: for any
$A^\star\in\mathbb{R}^{d\times d}$ and any vector $\mathbf{v}\in\mathbb{R}^d$,
\begin{equation}
\label{eq:singular_value_ineq}
\| \mathbf{v} A^\star \|_2 \;\geq\; \sigma_{\min}(A^\star)\,\|\mathbf{v}\|_2 ,
\end{equation}
where $\sigma_{\min}(A^\star)$ is the smallest singular value of $A^\star$.
Applying~\eqref{eq:singular_value_ineq} with $\mathbf{v}=\Delta\mathbf{z}_i$,
\[
\|\Delta\mathbf{z}_i A^\star\|_2
\;\geq\;
\sigma_{\min}(A^\star)\,\|\Delta\mathbf{z}_i\|_2 .
\]

Combining this with~\eqref{eq:triangle_step} yields
\[
\|\Delta\mathbf{g}_i\|_2
\;\geq\;
\sigma_{\min}(A^\star)\,\|\Delta\mathbf{z}_i\|_2 - \varepsilon_E,
\]
which is exactly~\eqref{eq:lb-l2-formal}.

\paragraph{Step 3: From $\ell_2$ to $\ell_1$ norms.}
We now derive a corresponding bound in $\ell_1$.
First, note that for any $\mathbf{v}\in\mathbb{R}^d$,
\begin{equation}
\label{eq:l1_l2_basic}
\|\mathbf{v}\|_2 \;\leq\; \|\mathbf{v}\|_1 ,
\end{equation}
and Hölder's inequality gives
\begin{equation}
\label{eq:holder}
\|\mathbf{v}\|_1 \;\leq\; \sqrt{d}\,\|\mathbf{v}\|_2
\quad\Longrightarrow\quad
\|\mathbf{v}\|_2 \;\geq\; \frac{1}{\sqrt{d}}\,\|\mathbf{v}\|_1 .
\end{equation}

Starting from~\eqref{eq:lb-l2-formal} and using~\eqref{eq:l1_l2_basic} on the left
and~\eqref{eq:holder} on the right, we obtain
\begin{align*}
\|\Delta\mathbf{g}_i\|_1
&\geq \|\Delta\mathbf{g}_i\|_2 \\
&\geq \sigma_{\min}(A^\star)\,\|\Delta\mathbf{z}_i\|_2 - \varepsilon_E \\
&\geq \sigma_{\min}(A^\star)\,\frac{1}{\sqrt{d}}\,\|\Delta\mathbf{z}_i\|_1
     - \varepsilon_E .
\end{align*}
This is precisely the claimed inequality~\eqref{eq:lb-l1-formal}.

Thus both bounds~\eqref{eq:lb-l2-formal} and~\eqref{eq:lb-l1-formal} hold, completing the proof.

                           

\section{Implementation Details}
\label{sec:im}
\noindent\textbf{Implementation details.}
In all experiments, we optimize a square noise patch of size \(50\times 50\) pixels placed on RGB observations of size \(224\times 224\).
The batch size is fixed to 2.
For the perturbation-augmentation stage, we set the budget on the sample-wise noise to \(\epsilon_{\sigma}=2/255\), and adopt a nested optimization with \(I=8\) inner steps and \(K=50\) outer steps.
The step sizes are \(\eta_{\sigma}=1/510\) for the sample-wise perturbations and \(\eta_{\delta}=1\times 10^{-3}\) for the universal patch. For different values of $\epsilon_\sigma$, the step size $\eta_\sigma$ is set such that $I \times \eta_\sigma=2 \times \epsilon_\sigma$.
The three loss components are weighted by \(\lambda_{\mathrm{con}}=10\), \(\lambda_{\mathrm{PAD}}=1\), and \(\lambda_{\mathrm{PSM}}=0.5\), respectively.
We run the optimization for 2000 iterations in all settings and report the performance at the final iteration.

For the InfoNCE loss, we use a temperature \(\tau=0.07\).
For the Patch Attention Dominance (PAD) term, we aggregate the last two text\(\to\)vision attention layers, apply a non-patch weight of \(\lambda_{\mathrm{non}}=0.8\), and restrict the attention reweighting to the top-\(\rho=0.3\) text tokens ranked by their clean attention mass.
We further enforce a margin constraint such that the patch-induced attention increment exceeds the strongest non-patch increment by at least \(m=0.1\).
For the Patch Semantic Misalignment (PSM) loss, we set \(\alpha=1.0\), \(\beta=0.5\), and use temperature \(\tau=0.3\) in the soft alignment terms.
The sensitivity of our method to these hyperparameters is analyzed in Appendix~\ref{sec:dab}.

\section{Main Results under White-box Setting}
\label{sec:white}
\begin{table*}[!t]
\centering
\small
\caption{We report the success rate (SR) on LIBERO simulation in a white-box setup. $^*$ marks an in-domain dataset matching the patch-training data, and $^\triangle$ marks a transfer evaluation on a different victim dataset. }
\resizebox{\linewidth}{!}{%
\begin{tabular}{l||ccccc|ccccc}
\toprule
\multirow{2}{*}{Objective} &
\multicolumn{5}{|c}{Simulated} &
\multicolumn{5}{|c}{Physical} \\
\cmidrule(lr){2-6}\cmidrule(lr){7-11}
& Spatial$^\triangle$ & Object$^\triangle$ & Goal$^\triangle$ & Long$^\ast$ & Avg.
& Spatial$^\triangle$ & Object$^\triangle$ & Goal$^\triangle$ & Long$^\triangle$ & Avg. \\
\midrule
Benign
& $84.7\pm10.2$ & $88.4\pm10.0$ & $79.2\pm12.0$ & $53.7\pm18.6$ & $76.5$
& $84.7\pm10.2$ & $88.4\pm10.0$ & $79.2\pm12.0$ & $53.7\pm18.6$ & $76.5$ \\
Random Noise
& $71.2\pm24.2$ & $85.2\pm7.9$ & $79.0\pm15.5$ & $51.6\pm14.8$ & $71.7$
& $71.2\pm24.2$ & $85.2\pm7.9$ & $79.0\pm15.5$ & $51.6\pm14.8$ & $71.7$ \\
\midrule
$\text{UMA}_{1}$
& $0.0\pm0.0$ & $1.0\pm3.0$ & $0.0\pm0.0$ & $0.0\pm0.0$ & $0.2$
& $10.4\pm15.3$ & $39.2\pm16.8$ & $35.2\pm26.8$ & $20.0\pm20.8$ & $26.2$ \\
$\text{UMA}_{1-3}$
& $0.0\pm0.0$ & $1.2\pm2.6$ & $1.0\pm2.4$ & $0.0\pm0.0$ & $0.5$
& $3.4\pm6.2$ & $43.6\pm20.4$ & $20.0\pm20.5$ & $18.0\pm17.9$ & $21.2$ \\
$\text{UADA}_{1}$
& $0.0\pm0.0$ & $0.8\pm2.4$ & $0.0\pm0.0$ & $0.0\pm0.0$ & $0.2$
& $0.8\pm1.8$ & $1.2\pm2.4$ & $7.4\pm15.2$ & $3.4\pm4.8$ & $3.2$ \\
$\text{UADA}_{1-3}$
& $0.0\pm0.0$ & $0.0\pm0.0$ & $0.0\pm0.0$ & $0.0\pm0.0$ & $0.0$
& $0.0\pm0.0$ & $0.0\pm0.0$ & $0.0\pm0.0$ & $0.0\pm0.0$ & $0.0$ \\
UPA
& $3.8\pm11.4$ & $22.2\pm12.5$ & $12.0\pm10.4$ & $3.2\pm3.0$ & $10.3$
& $4.4\pm9.5$ & $43.0\pm23.6$ & $42.8\pm24.4$ & $27.2\pm19.7$ & $29.3$ \\
TMA (Avg.)
& $0.8$ & $13.6$ & $11.1$ & $2.0$ & $6.9$
& $9.9$ & $48.9$ & $41.3$ & $21.1$ & $30.3$ \\ \midrule
\textbf{Our }& $0.0\pm0.0$ & $0.0\pm0.0$ & $2.0\pm0.0$ & $0.0\pm0.0$ & $0.5$
& $0.0\pm0.0$ & $9.0\pm0.0$ & $0.0\pm0.0$ & $2.0\pm0.0$ & $2.75$ \\
\bottomrule
\end{tabular}
}%
\label{tb:white}
\end{table*}

Tab.~\ref{tb:white} evaluates task success rates in the LIBERO simulator under a white-box setup.
Although our objective is explicitly designed for black-box transfer, it still shows competitive white-box performance. In the simulated setting, our patch almost completely disables the policy, driving success rates to near zero across all suites with an average of only 0.5\%, on par with the strongest UMA/UADA variants and far below UPA and TMA (10.3\% and 6.9\% on average). In the physical setting, our method again reduces success to almost zero (2.75\% on average), ranking second only to UADA$_{1-3}$ and clearly outperforming UPA and TMA. These results indicate that the proposed universal patch retains strong white-box attack capability while being tailored for transfer.

\section{Main Results on $\pi_0$}
\label{sec:pi0}

\begin{table*}[t]
\centering
\caption{Task success rate (\%) when transfer from the surrogate OpenVLA-7B to the victim $\pi_0$ on LIBERO. }
\begin{tabular}{c||ccccc|ccccc}
\toprule
\multirow{2}{*}{objective} & \multicolumn{5}{c|}{Simulated}        & \multicolumn{5}{c}{Physical}          \\ \cmidrule{2-11}
                           & spatial & object & goal & long & avg. & spatial & object & goal & long & avg. \\ \midrule
Benign & 96&	98&	95&	79&	92.00& 96&	98&	95&	79&	92.00 \\
$\text{UMA}_1$& 100 & \textbf{94}  & 91 &  \textbf{72}& 89.25 & 98 & 99 & 98 & 79 &93.50 \\
$\text{UMA}_{1-3}$  &  99 &  97 & 95 & 77& 92.00 & 97  & 97 & 90 & 72& 89.00     \\
$\text{UADA}_{1}$    &  93& 96 &90 &75  & 88.50 & \textbf{93}& 94 & 96 & 73 & 89.00 \\ 
$\text{UADA}_{1-3}$    &95 & 96 & 96  & 79& 91.50& 96 & 96 &  94  & 70 & 89.00 \\  
$\text{DOF}_1$ & 98 &97  &90 & 72 & 89.25& 96  &97 &94 & 78 & 91.25 \\ 
$\text{DOF}_7$ &  98& 97  & 94 & 75 &91.00 & 97  & 99 & 86 &79 & 90.25 \\ 
\midrule
Our  &\textbf{91}&96&\textbf{85}&\textbf{72}&\textbf{86.00}& \textbf{93} &\textbf{92} & \textbf{82} & \textbf{67} & \textbf{83.50} \\
\bottomrule               
\end{tabular}
\label{tb:pi0}
\end{table*}

In the main text we reported transfer results from \emph{OpenVLA-7B} \citep{kim2024openvla} to \emph{OpenVLA-oft-w} \citep{kim2025fine} and \emph{OpenVLA-oft}.
Tab.~\ref{tb:pi0} complements these experiments by showing transfer to the $\pi_0$~\citep{black2024pi_0}. This transfer is substantially harder, since $\pi_0$ differs from OpenVLA along almost every axis, including model architecture, pretraining pipeline, training data, and action head design, making cross-model transfer particularly challenging. 

Even under this large surrogate to victim gap, our universal patch still achieves the strongest degradation in task success. In the simulated setting, the benign policy succeeds on 92.0\% of tasks on average, whereas our method reduces the success rate to 86.0\%, which is 2.5\% percentage points lower than the best baseline (UADA$_1$, 88.5\%). The advantage becomes even clearer in the physical setting: our average success rate of 83.50\% is 5.50\% points below the strongest baseline (89.0\%), while other objectives stay closer to the benign performance. These results indicate that our feature and attention level design remains effective even when transferring from OpenVLA-7B to a structurally and procedurally very different VLA model, and highlight our superior transferability under the challenging physical to simulation cross-setting transfer. 

\begin{table}[t]
    \centering
    \caption{Ablation on patch size for transfer to openvla-oft in the physical setting.}
\begin{tabular}{c|ccccc}
\toprule
Patch size &  spatial & object & goal & long & avg. \\ \midrule
3\% & 79& 86& 88& 66& 79.75\\
5\% & 69 &74 & 76 & 27 & 61.50\\
7\% &28&  78&35&15& 39.00\\
10\%&41 &6 &35&1&20.75\\
\bottomrule
\end{tabular}
\label{tab:ps}
\end{table}


\section{Detailed Ablation Study}
\label{sec:dab}

\begin{table}[t]
    \centering
    \caption{Ablation on $\lambda_\mathrm{con}$ for transfer to openvla-oft in the physical setting.}
\begin{tabular}{c|ccccc}
\toprule
objective &  spatial & object & goal & long & avg. \\ \midrule
$\lambda_\mathrm{con}=1$ &68 &77& 75& 35&63.75 \\
$\lambda_\mathrm{con}=2$ & 72 &73 & 76&28&62.25\\
$\lambda_\mathrm{con}=5$ & 70 &77 &67&33& 61.75\\
$\lambda_\mathrm{con}=10$ & 69 &74 & 76 & 27 & 61.50\\
\bottomrule
\end{tabular}
\label{tab:con}
\end{table}

\begin{table}[]
    \centering
    \caption{Ablation on $\epsilon$ in RUPA for transfer to openvla-oft in the physical setting.}
\begin{tabular}{c|ccccc}
\toprule
objective &  spatial & object & goal & long & avg. \\ \midrule
$\epsilon=1/255$ &73 & 73 & 77 & 28 & 62.75 \\
$\epsilon=2/255$ & 69 &74 & 76 & 27 & 61.50 \\
$\epsilon=4/255$ & 66 & 71 & 62 & 33 & 58.00 \\
$\epsilon=8/255$ & 72 & 62 & 70 &38& 60.50\\
$\epsilon=16/255$ & 75& 67 & 68 & 36 & 61.50\\
\bottomrule
\end{tabular}
\label{tab:eps}
\end{table}

\noindent\textbf{Impact of Patch Size.}
Tab.~\ref{tab:ps} ablates the patch area, varying it from 3\% to 10\% of the input image. We observe a clear monotonic trend: larger patches yield stronger attacks. A very small 3\% patch already degrades performance compared to the baseline methods, but still leaves a high average success rate of 79.75\%, indicating limited capacity to transfer attack. Increasing the size to 5\%, our default choice, substantially strengthens the attack, reducing the average success rate to 61.50\% while keeping the patch relatively compact and unobtrusive. When the patch occupies 7\% or 10\% of the image, the policy is almost completely disabled (39.00\% and 20.75\% on average), with object-centric success even dropping to 6\% at 10\%. This suggests that once the patch area is large enough to consistently intersect action-relevant regions, our feature and attention based objectives can fully dominate the visual stream. In practice, 5\% offers a favorable trade-off between visual footprint and attack strength, while larger patches mainly amplify the effect rather than changing the attack behavior.


\noindent\textbf{Impact of $\lambda_\mathrm{con}$.}
Tab.~\ref{tab:con} ablates the weight $\lambda_\mathrm{con}$ that balances the feature-space $\ell_1$ term and the contrastive loss in our objective (both coefficients are rescaled by a factor of 0.1 during optimization for numerical stability). We observe that increasing $\lambda_\mathrm{con}$ from 1 to 10 steadily strengthens the attack, with the average success rate dropping from 63.75\% to 61.50\% and saturating once $\lambda_\mathrm{con} \ge5$. This trend is consistent with Tab.~\ref{tab:ab} and our theory that $\mathcal{L}_\mathrm{con}$ primarily controls the \emph{direction} of feature displacement, while $\mathcal{L}_1$ controls its magnitude: when $\lambda_\mathrm{con}$ is too small, the $\ell_1$ term dominates and the patch mainly enlarges deviations without steering them into transferable directions; giving the contrastive term comparable or larger weight leads to more aligned, high-CCA feature shifts and thus better cross-model transfer. At the same time, the plateau between $\lambda_\mathrm{con}=5$ and 10 indicates that our method is not overly sensitive once the contrastive component is sufficiently emphasized.

\noindent\textbf{Impact of $\epsilon$ in RUPA.}
Tab.~\ref{tab:eps} ablates the perturbation bound $\epsilon$ used for sample-wise inner minimization in Phase 1 of RUPA. Recall that these per-sample perturbations act as on-the-fly “hard” augmenters around each patched input. We observe that moderate noise levels yield the strongest transfer: increasing $\epsilon$ from 1 to 4 steadily lowers the average success rate from 62.75\% to 58.00\%, while further enlarging $\epsilon$ to 8 or 16 degrades performance again (60.5\% and 61.5\%).

This pattern suggests that RUPA behaves like a localized adversarial training loop around the universal patch. When $\epsilon$ is too small, the inner minimization explores only a narrow neighborhood and fails to expose the patch to sufficiently challenging geometric and appearance variations, limiting robustness. A moderate $\epsilon$ ($\epsilon=4/255$) encourages the patch to align with features that remain effective within a realistic but nontrivial perturbation ball, leading to better transfer. However, overly large $\epsilon$ pushes samples far from the natural data manifold; the inner loop then overfits to unrealistic, heavily corrupted views, which weakens the invariances shared between surrogate and victim and ultimately harms black-box performance.

\begin{figure*}[t]
    \centering
    \includegraphics[width=0.12\linewidth]{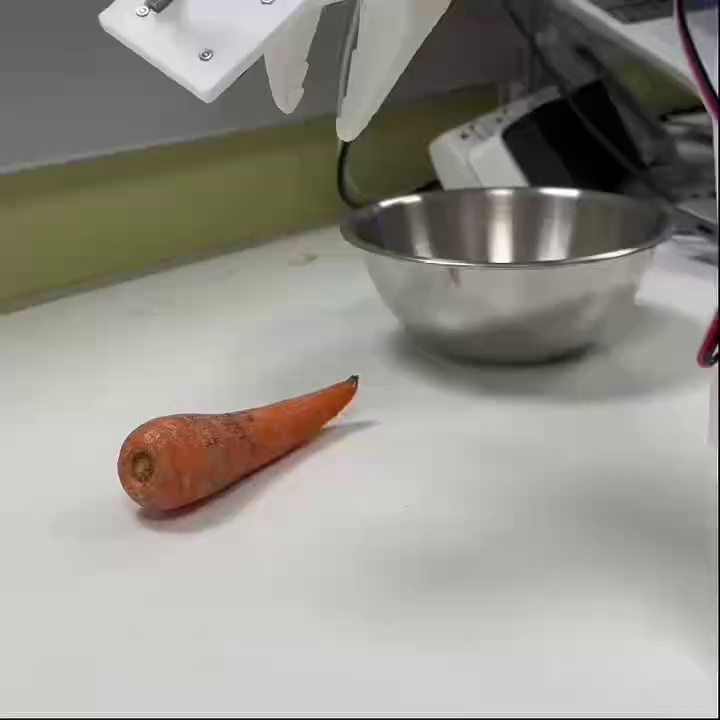}
    \includegraphics[width=0.12\linewidth]{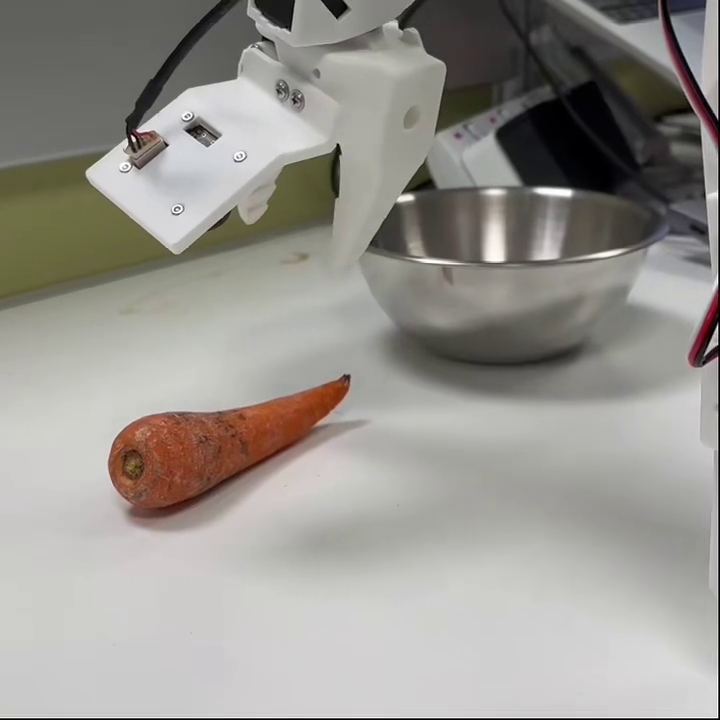}
    \includegraphics[width=0.12\linewidth]{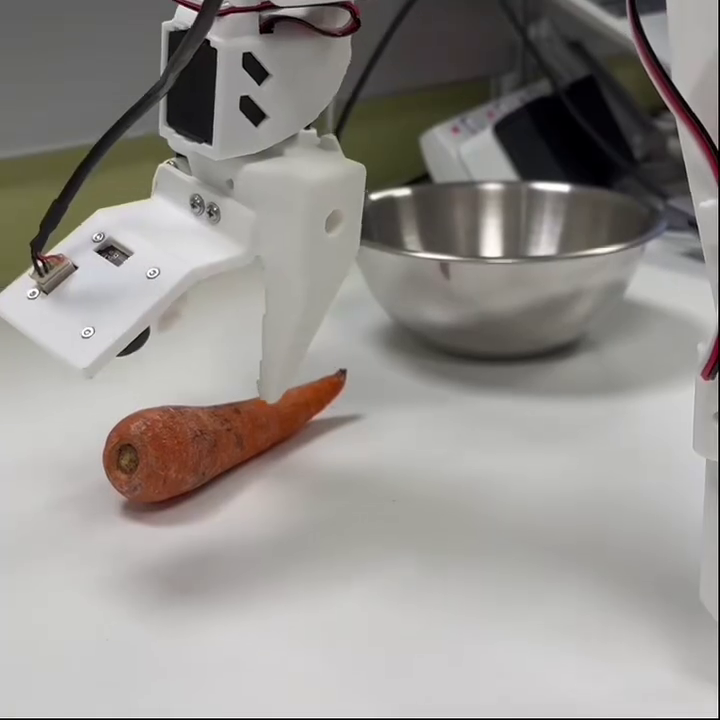}
    \includegraphics[width=0.12\linewidth]{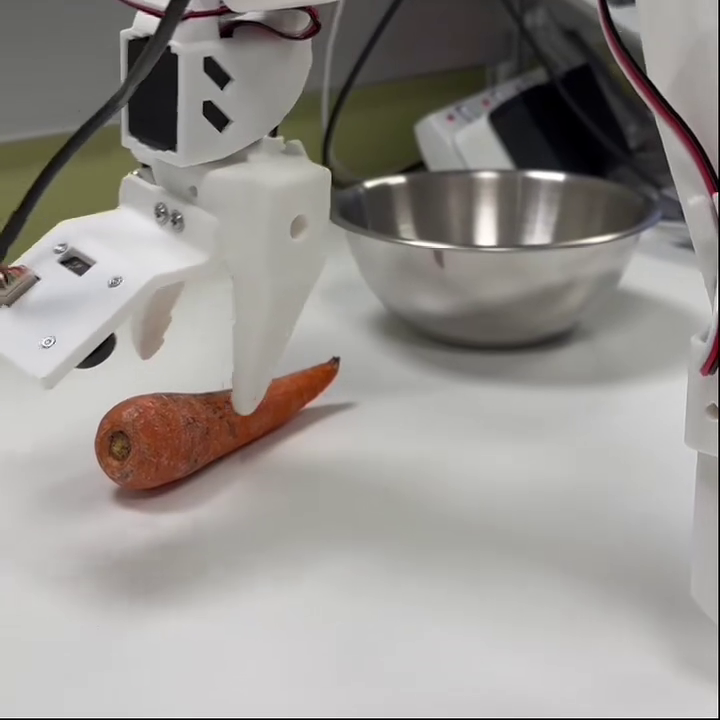}
    \includegraphics[width=0.12\linewidth]{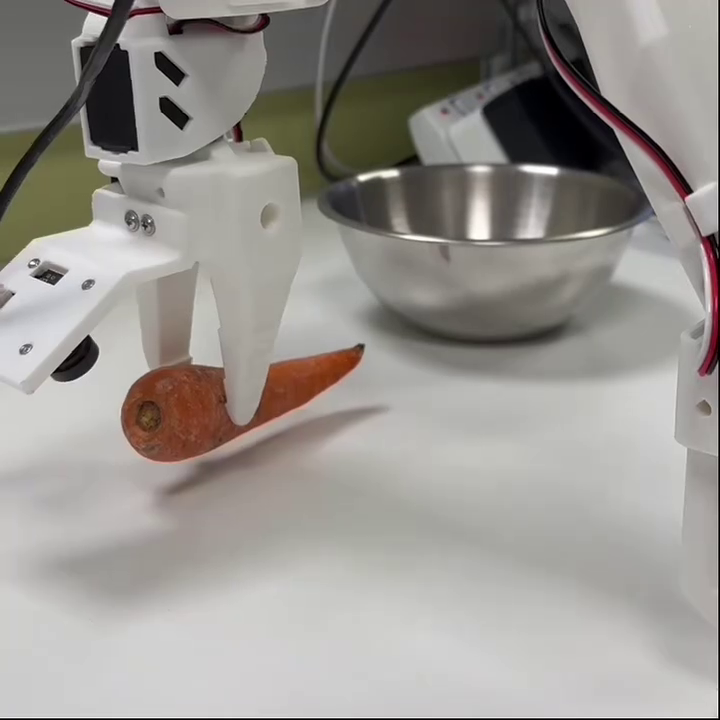}
    \includegraphics[width=0.12\linewidth]{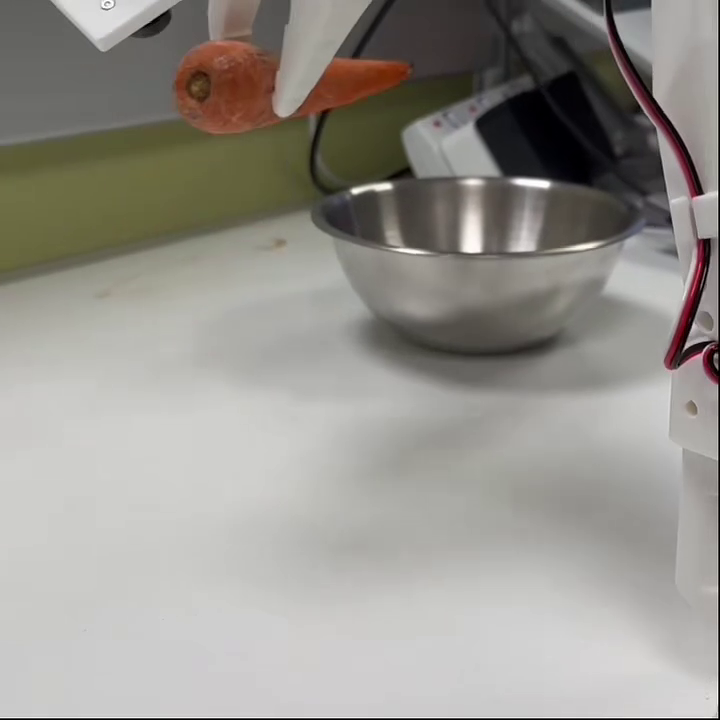}
    \includegraphics[width=0.12\linewidth]{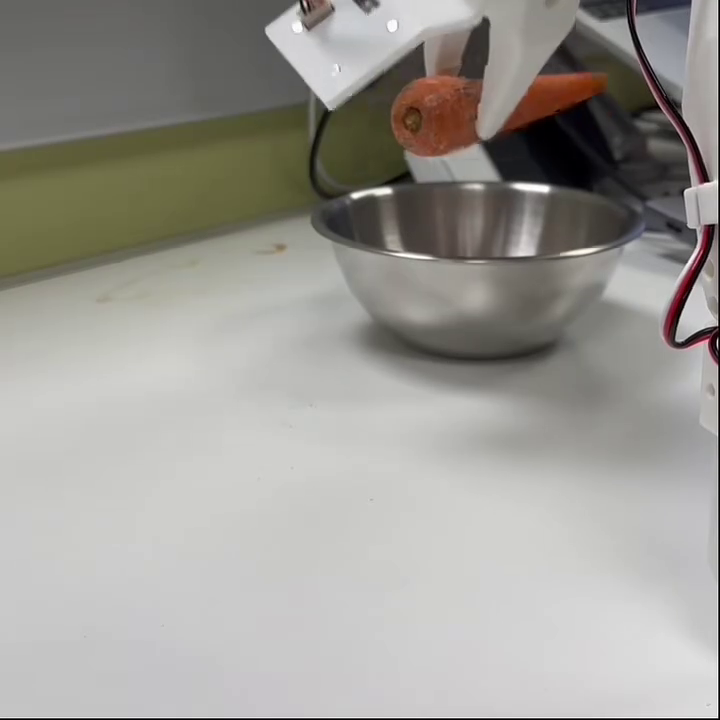}
    \includegraphics[width=0.12\linewidth]{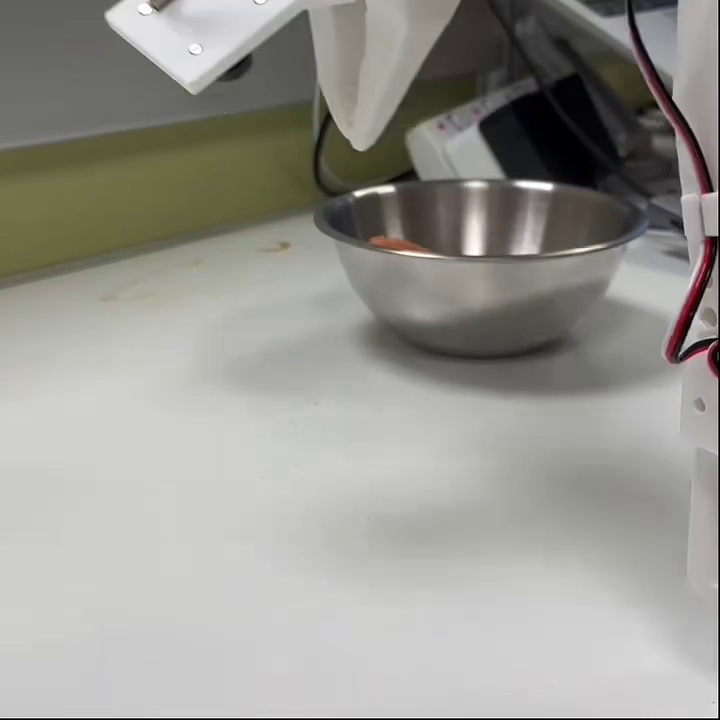}
    \includegraphics[width=0.12\linewidth]{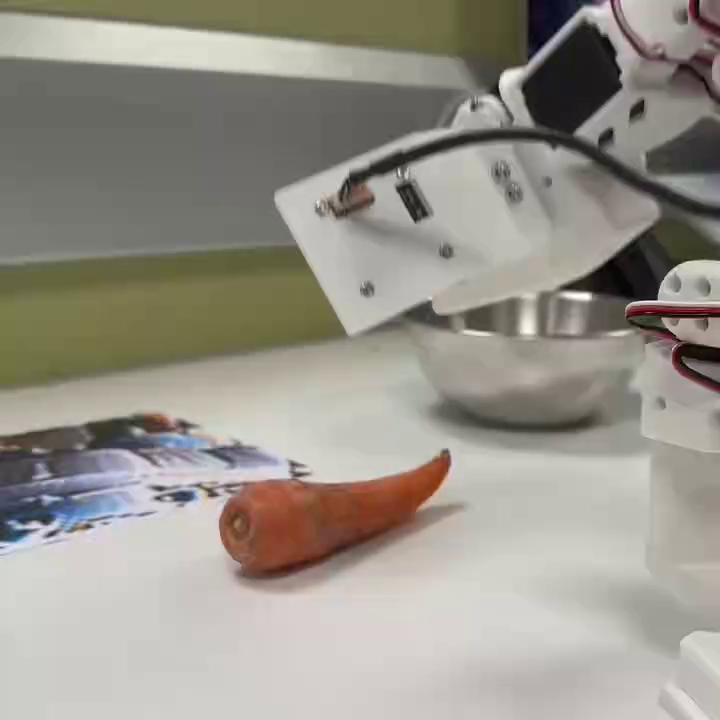}
    \includegraphics[width=0.12\linewidth]{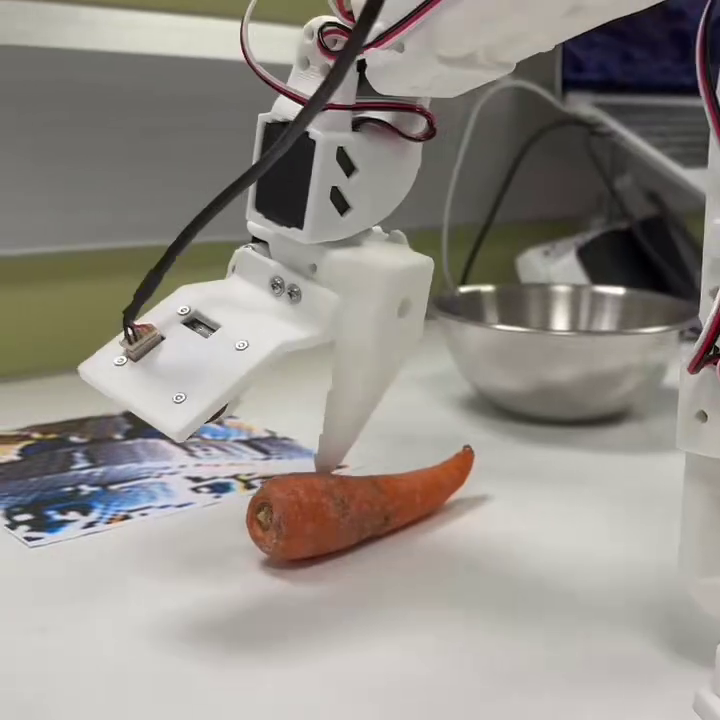}
    \includegraphics[width=0.12\linewidth]{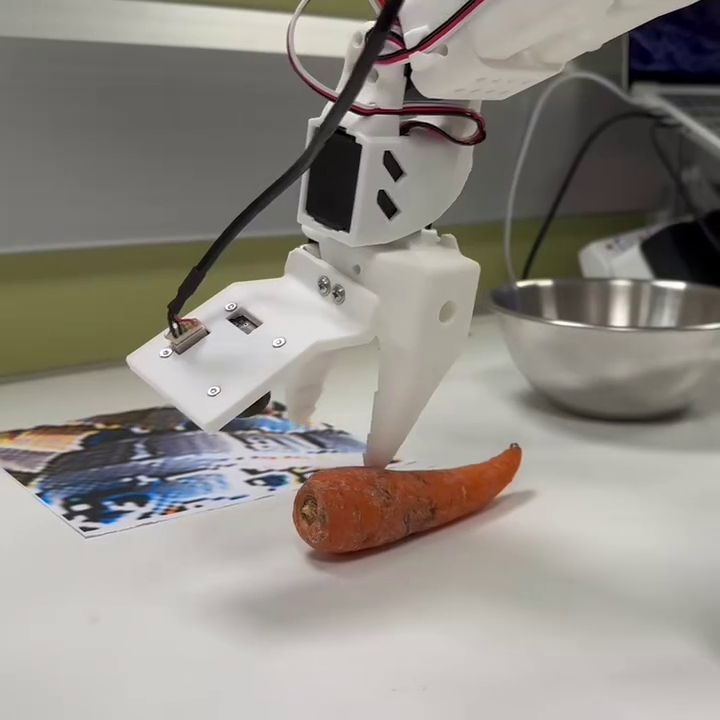}
    \includegraphics[width=0.12\linewidth]{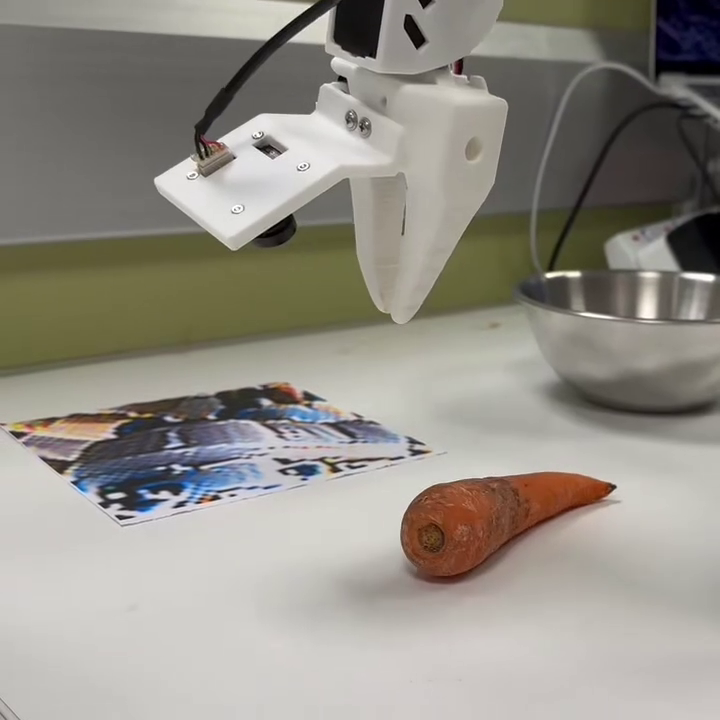}
    \includegraphics[width=0.12\linewidth]{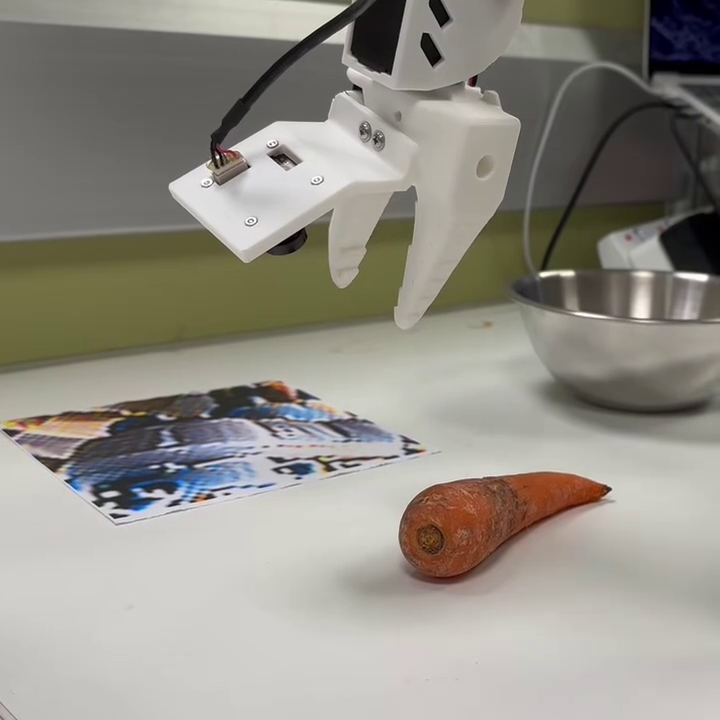}
    \includegraphics[width=0.12\linewidth]{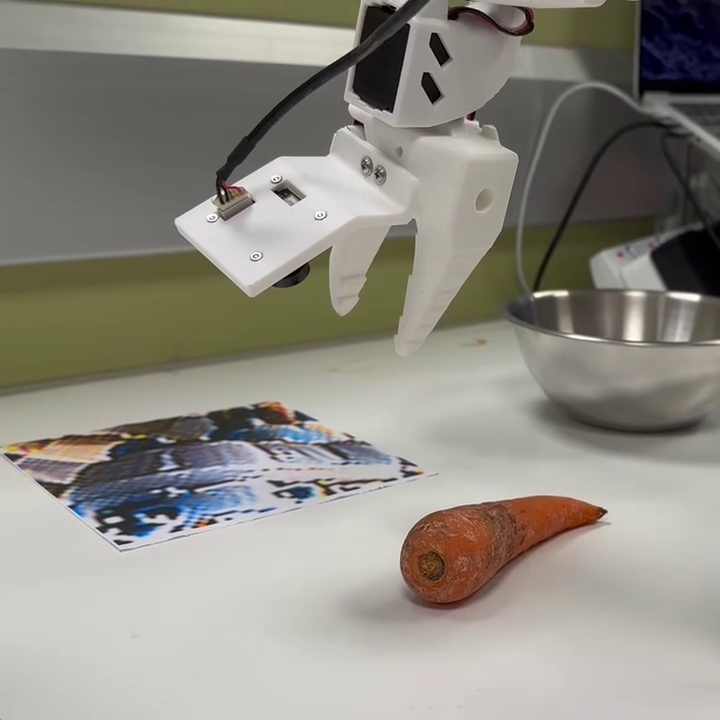}
    \includegraphics[width=0.12\linewidth]{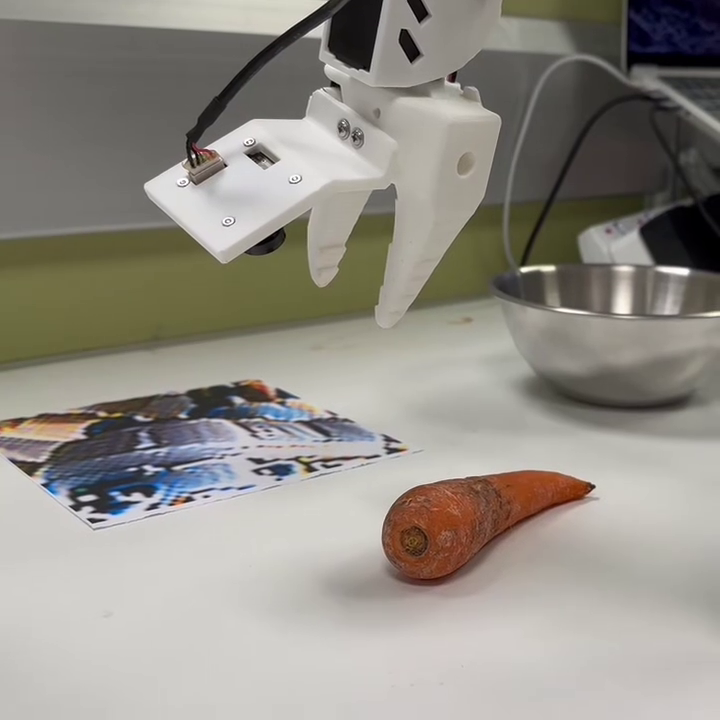}
    \includegraphics[width=0.12\linewidth]{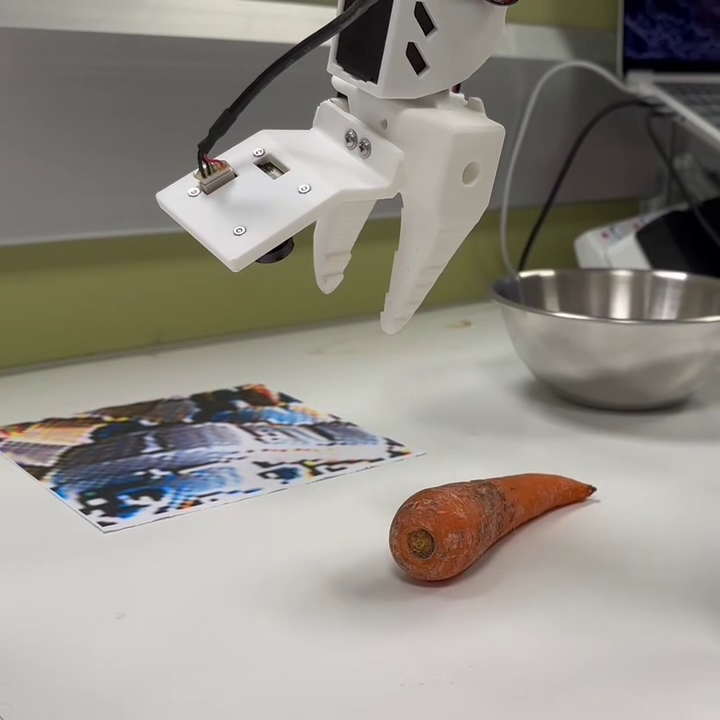}
    \caption{\textbf{Qualitative real-world results.} The top row displays benign executions, while the bottom row shows their adversarial counterparts.}
\label{fig:real}
\end{figure*}
\begin{figure*}[t]
    \centering
    \includegraphics[width=0.12\linewidth]{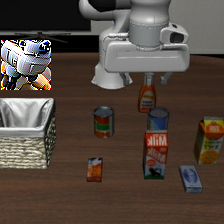}
    \includegraphics[width=0.12\linewidth]{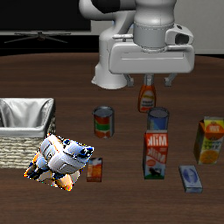}
    \includegraphics[width=0.12\linewidth]{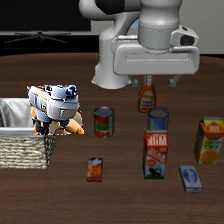}
    \includegraphics[width=0.12\linewidth]{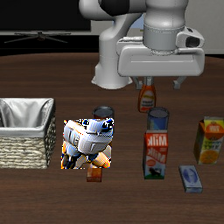}
    \includegraphics[width=0.12\linewidth]{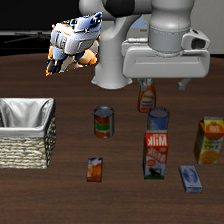}
    \includegraphics[width=0.12\linewidth]{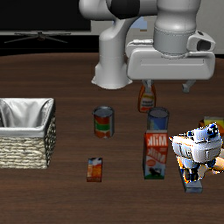}
    \includegraphics[width=0.12\linewidth]{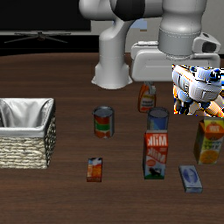}
    \includegraphics[width=0.12\linewidth]{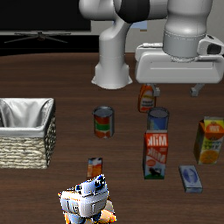}
    \includegraphics[width=0.12\linewidth]{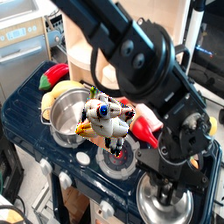}
    \includegraphics[width=0.12\linewidth]{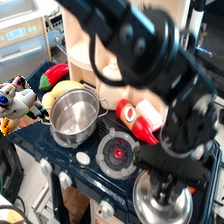}
    \includegraphics[width=0.12\linewidth]{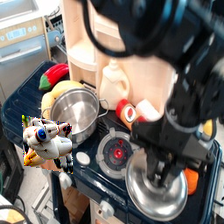}
    \includegraphics[width=0.12\linewidth]{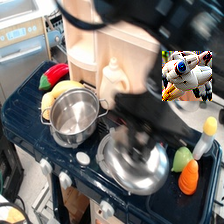}
    \includegraphics[width=0.12\linewidth]{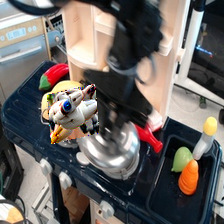}
    \includegraphics[width=0.12\linewidth]{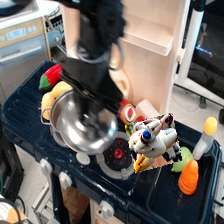}
    \includegraphics[width=0.12\linewidth]{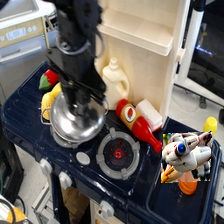}
    \includegraphics[width=0.12\linewidth]{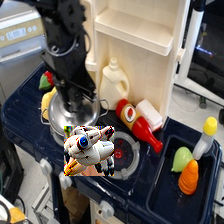}
    \caption{\textbf{Training videos from simulated and physical settings.} The top row shows eight frames sampled from a simulated training video, while the bottom row shows eight frames from a physical training video.}
\label{fig:train}
\end{figure*}

\section{Real-world Performance}
Beyond digital simulation, we qualitatively assess our adversarial patches in a physical robot setup under a black-box setting. We run repeated trials across three distinct tasks, including object grasping, placement, and manipulation, 3 times. As shown in Fig.~\ref{fig:real}, the patch reliably steers the robot to fail all tested executions. In the real world, each task failure represents a successful transfer attack on the black-box VLA model, highlighting the strong real-world transferability of our method. Detailed recordings are provided as videos in the supplementary material. From the videos, we observe that the attack is insensitive to patch location: across three qualitative trials, patches placed at different positions consistently cause the tasks to fail.

\section{Training Video Visualisation}
Figure~\ref{fig:train} illustrates the training videos used for our universal patch optimization. The top row shows eight frames from a simulated setting, and the bottom row shows eight frames from a physical setting. In both rows, frames include sample-wise perturbations and patch geometric transformations (random position, skew, and rotation). The sample-wise perturbations are bounded by $\epsilon=2/255$, making them imperceptible to the human eye and thus unlikely to affect real-world test performance. The patch geometric transformations follow the implementation of RoboticAttack~\citep{wang2025exploring}. Additional qualitative comparisons between our patch and the baseline patch on LIBERO are provided as videos in the supplementary material.

\noindent\textbf{Why is physical transfer harder?}
Fig.~\ref{fig:train} highlights a pronounced gap between simulated and physical training videos: the physical scenes exhibit richer clutter, stronger noise and motion blur, and more severe perspective distortions, leading to a much broader and more complex perceptual distribution. In simulation, actions are almost directly driven by visual tokens, so misguiding them quickly causes failure. Whereas on the real robot, trajectory smoothing and mechanical redundancy can partially compensate for perturbed decisions. These factors together make cross-setting transfer substantially harder and explain the larger performance gap between simulated and physical attacks.


\end{document}